\documentclass[lic,lith,english]{liuthesis}

\usepackage{multirow}
\usepackage{listings}
\usepackage{xcolor}

\lstdefinestyle{mypython}{
	language=Python,
	basicstyle=\ttfamily\small,
	numbers=left,
	numberstyle=\tiny,
	stepnumber=1,
	numbersep=5pt,
	showstringspaces=false,
	tabsize=4,
	breaklines=true,
	frame=single,
	captionpos=b
}

\department{Department of Computer and Information Science}
\departmentenglish{STIMA}
\departmentshort{IDA}
\edition{1:1}

\supervisor{Dong Qian}
\examiner{Zheng Zhao}

\titleenglish{Discrete Diffusion Models for Language Generation}
\titleswedish{Diskreta diffusionsmodeller för språkgenerering}
\subtitleenglish{Diskreta diffusionsmodeller för språkgenerering}

\division{Division of Human-Centered Systems}
\thesisnumber{25/008}
\publicationyear{2025}

\isbnprint{LIU-IDA}{STAT-A--25}{008--SE}
\isbnpdf{LIU-IDA}{STAT-A--25}{008--SE}

\author{Thalgamuwe Gedara Ashen Akalanka Weligalle}

\begin{document}
	




\ifxetex 
\fi


\newif\ifNoChapNumber
\newcommand\Vlines{%
\def\VL{\rule[-2cm]{1pt}{5cm}\hspace{1mm}\relax}
\VL\VL\VL\VL\VL\VL\VL}
\makeatletter
\setlength\midchapskip{0pt}
\makechapterstyle{VZ43}{
\renewcommand\chapternamenum{}
\renewcommand\printchaptername{}
\renewcommand\printchapternum{}

\renewcommand\chapnumfont{\Huge\bfseries\centering}
\renewcommand\chaptitlefont{\Huge\bfseries\raggedright}
\renewcommand\printchaptertitle[1]{%
\Vlines\hspace*{-2em}%
\begin{tabular}{@{}p{1cm} p{\textwidth-3cm}}%
\ifNoChapNumber\relax\else%
\colorbox{black}{\color{white}%
\makebox[.8cm]{\chapnumfont\strut \thechapter}}
\fi
& \chaptitlefont ##1
\end{tabular}
\NoChapNumberfalse
}
\renewcommand\printchapternonum{\NoChapNumbertrue}
}
\makeatother


\setcounter{secnumdepth}{3} 
\setcounter{tocdepth}{1} 







	\chapterstyle{VZ43}
	

\chapter{Introduction}
\label{cha:introduction}

The remarkable progress in generative AI has opened new frontiers in machine learning, particularly in domains such as image synthesis and video generation. From hyper-realistic artwork generated by diffusion models to photorealistic deepfakes, these advancements have significantly reshaped our expectations of artificial intelligence. While diffusion models have shown state-of-the-art performance in modeling continuous data, their applicability to discrete domains—such as natural language—remains a compelling area of research.

Natural language generation has historically been dominated by autoregressive (AR) models \ref{sec:ar} such as GPT-2\footnote{\url{https://huggingface.co/openai-community/gpt2}}, which generate text sequentially by predicting one token at a time. Although highly effective, AR models have known limitations, including exposure bias and limited support for parallel computation during inference. 

As a compelling alternative, recent research has explored diffusion-based approaches for discrete data generation \cite{sahoo2024simple}\cite{ho2020denoising}\cite{austin2021structured}. These models operate by progressively refining a noisy input sequence through a series of denoising steps, allowing them to model complex distributions in a non-autoregressive manner. Unlike AR models, diffusion models generate entire sequences in parallel during inference, offering potential advantages in sampling flexibility, controllability, and robustness against exposure bias. 

This thesis aims to investigate the potential of discrete diffusion models in the context of natural language generation. In particular, we focus on the Discrete Denoising Diffusion Probabilistic Model (D3PM)\footnote{\url{https://github.com/kuleshov-group/mdlm}}, a variant of diffusion models tailored for discrete data. By comparing the performance of D3PM with that of established autoregressive models such as PixelSNAIL~\cite{chen2018pixelsnail}, we aim to identify the strengths, limitations, and areas for improvement of each modeling paradigm.

Unlike autoregressive models, which process and predict tokens sequentially, discrete diffusion models operate in a fundamentally different way. They employ a multi-step noising~\ref{d3pm:forward} and denoising process~\ref{d3pm:backward} and are inherently parallel, potentially offering advantages in both efficiency and robustness. This thesis seeks to answer whether diffusion-based approaches can not only match but potentially replace autoregressive models in certain contexts.

Both model families are built on the transformer architecture—AR~\ref{ar:architecture} models using transformer decoders, and diffusion models employing transformers as denoisers. However, the fundamental distinction lies in their operational mechanics: while autoregressive models predict the next token in a sequence, diffusion models iteratively mask and unmask data over a series of steps.

The introduction can be divided into these sections:

\section{Motivation}
\label{sec:motivation}

Numerous research efforts have explored the comparative performance of text generation models, focusing on diverse architectures and objectives. For instance, studies have examined Multimodal Models versus Permutation-based Models~\cite{hirt2023learning}, Mixture-of-Experts compared to Multimodal approaches~\cite{yu2023mmoe}, and Encoder-Decoder Transformers versus Multimodal Transformers~\cite{qi2022medt}. These comparisons are often context-dependent, targeting specific challenges such as multimodal integration, generation coherence, or scalability.

This research is primarily motivated by the need to compare two distinct families of language models: Autoregressive (AR) models and Discrete Diffusion models. AR models, such as GPT-2, generate text sequentially and have long been considered the state-of-the-art in natural language generation. However, they suffer from well-known limitations, including exposure bias and restricted support for parallel computation during inference.

In contrast, Discrete Diffusion models, such as the masked Discrete Denoising Diffusion Probabilistic Model (D3PM)~\cite{austin2021structured}, offer a fundamentally different approach. These models perform generation by learning to reverse a noising process applied to discrete tokens. One key advantage of diffusion models is their ability to generate tokens in parallel, potentially addressing some of the core inefficiencies inherent in AR architectures.

This research is driven by the following guiding questions:
\begin{itemize}
    \item Can discrete diffusion models achieve competitive performance in language generation tasks traditionally dominated by autoregressive models?
    \item What are the practical trade-offs in terms of generation quality, efficiency, and training complexity?
    \item Are there specific use cases where diffusion models are better suited than AR models?
\end{itemize}

To explore these questions, we conduct a systematic comparative analysis of AR and discrete diffusion models under a unified evaluation framework. All models are trained and tested under identical conditions to ensure fairness and reproducibility. Performance is assessed using standard metrics such as Bits Per Token (BPT)\ref{bpt:bpc to bpc}, Negative Log-Likelihood (NLL)\ref{sec:nll}, Perplexity (PPL)\ref{sec:ppl}, and generation speed.

\section{Aim}
\label{sec:aim}

The primary objective of this study is to conduct a comprehensive comparative analysis of discrete diffusion models for language generation, with a specific focus on their applicability in contrast to autoregressive (AR) models. This investigation is both statistical and structural, grounded in data-driven methodologies to evaluate performance across key metrics.

By systematically assessing the strengths, limitations, and trade-offs between discrete diffusion and AR architectures, this research seeks to provide actionable insights into the suitability of each model type for various language generation tasks. Through controlled experiments using standardized datasets and consistent evaluation conditions, the study aims to:

\begin{itemize}
    \item Analyze the generative efficiency and accuracy of discrete diffusion models.
    \item Determine whether these models can serve as practical alternatives to AR models in real-world applications.
    \item Contribute to the ongoing discourse on non-autoregressive generation in natural language processing.
\end{itemize}

Ultimately, the goal is to evaluate whether discrete diffusion techniques can offer scalable, robust, and high-quality alternatives to traditional autoregressive approaches in the domain of language modeling.

\section{Research questions}
\label{sec:research-questions}

This thesis aims to explore the effectiveness of discrete diffusion models, particularly D3PM and its simplified variants, in comparison to traditional autoregressive (AR) models in language generation tasks. The key research questions guiding this work are as follows:

\begin{itemize}
    \item How does the parallel token generation in D3PM contrast with the sequential decoding in autoregressive models?
    \item How do these methods perform in language generation tasks? What are the trade-offs, strengths, and limitations associated with each approach?
\end{itemize}

To address these questions, we follow a methodical approach that includes theoretical analysis, model implementation, and experimental evaluation. The answers to these questions will be developed throughout the subsequent sections of this thesis.





\chapter{Theory}
\label{cha:theory}

This chapter presents the theoretical foundations that underpin the models and evaluation techniques used in this research. It consolidates insights gathered from relevant academic literature and integrates them with the practical components of the thesis.

We begin by discussing the \textbf{Autoregressive Language Model} (Section~\ref{sec:ar}), followed by an exploration of the \textbf{Discrete Denoising Diffusion Probabilistic Model (D3PM)} (Section~\ref{sec:d3pm}), both of which serve as the central generative modeling approaches in this study.

Additionally, we elaborate on the statistical metrics employed to evaluate the performance and behavior of these models. These include:
\begin{itemize}
    \item \textbf{Bits Per Token (BPT)} (Section~\ref{sec:bpt}) – a measure of model efficiency and compression.
    \item \textbf{Perplexity (PPL)} (Section~\ref{sec:ppl}) – an indicator of the model's ability to predict unseen data.
    \item \textbf{Negative Log-Likelihood (NLL)} (Section~\ref{sec:nll}) – a loss-based evaluation metric derived from the log-likelihood of the predicted distribution.
\end{itemize}

Together, these theoretical and statistical components form the analytical backbone of the research and provide a framework for interpreting experimental results in subsequent chapters.

\section{Bits Per Token (BPT)}
\label{sec:bpt}

In the evaluation of language models, \textit{Bits Per Token} (BPT) is a fundamental metric that quantifies the average number of bits required to encode each token\footnote{A token is a basic unit of text, such as a word, subword, or character, used by language models to process and generate language.} in a given sequence. In this context, a token typically refers to a word or subword unit derived from a tokenized version of the dataset.

BPT provides valuable insight into a model’s predictive efficiency and its ability to compress linguistic information. A lower BPT value signifies that the model is more effective at assigning high probabilities to the correct tokens, thereby encoding information more compactly. This makes BPT particularly useful when evaluating the performance of generative language models, where efficient representation and accurate prediction are critical.

Overall, BPT not only reflects the model's compression capacity but also correlates strongly with its overall language modeling quality, making it an essential diagnostic in model comparison and benchmarking.

The concept of BPT is closely related to \emph{Bits Per Character} (BPC) \cite{blevins2019better}, as BPC provides a more granular perspective on information encoding at the character level. Therefore, to fully understand BPT, it is beneficial to first examine BPC in detail.

\subsection{Bits Per Character (BPC)}
\label{bpt:bpc}

\textit{Bits Per Character} (BPC) is a widely used metric for evaluating the efficiency of language models and data compression techniques. It quantifies the average number of bits required to represent each character within a dataset. A lower BPC value signifies more efficient encoding, indicating improved compression and predictive performance. This metric is particularly useful for assessing the effectiveness of language models in handling character-level representations of text.\cite{blevins2019better}

By understanding BPC, we can extend this analysis to BPT, where the unit of measurement shifts from individual characters to meaningful linguistic units, such as words or subwords. This transition is essential for evaluating modern language models, which often rely on subword tokenization techniques to balance efficiency and expressiveness.

The BPC for a given sequence of characters \( c \) is defined as:
$$BPC(c) = -\frac{1}{|c|} \sum_{i=1}^{|c|} \log_2 P(c_i | c_{<i})$$

where:
\begin{itemize}
    \item \( |c| \) is the total number of characters in the sequence.
    \item \( c_i \) is the \( i \)-th character in the sequence.
    \item \( c_{<i} \) represents all characters before \( c_i \) (i.e., the context for prediction).
    \item \( P(c_i | c_{<i}) \) is the probability assigned by the model to character \( c_i \), given the previous characters.
\end{itemize}

\subsection{BPC to BPT}
\label{bpt:bpc to bpc}

Mathematically, BPT is defined as the cross-entropy between the true data distribution and the model's predicted distribution over tokens:

$$BPT(W) = -\frac{1}{N} \sum_{i=1}^{N} \log_2 P(W_i | W_{<i})$$

where $N$ represents the total number of tokens, and $P(W_i | W_{<i})$ denotes the probability assigned by the model to the i-th token given its preceding context.

\section{Perplexity (PPL)}
\label{sec:ppl}

Perplexity \cite{fang2024wrong} is a widely used metric to evaluate a language model's ability to predict the next token in a sequence. Originally introduced by Jelinek et al. (1977) \cite{jelinek1977perplexity}, it measures how well a probability model predicts a sample.
Given a sequence of tokens $\mathbf{x} = (x_1, x_2, \dots, x_n)$ a language model parameterized by \( \theta \) estimates the conditional probability of each token given the preceding context:
\[
P_\theta(x_i \mid \mathbf{x}_{<i}), \quad i \in [1, n].
\]

The \textbf{perplexity (PPL)} \cite{fang2024wrong}over the sequence \( \mathbf{x} \) is defined as the inverse of the geometric mean of the predicted probabilities:
\[
\text{PPL}_\theta(\mathbf{x}) = \exp\left(-\frac{1}{n} \sum_{i=1}^{n} \log P_\theta(x_i \mid \mathbf{x}_{<i})\right).
\]

Equivalently, this can be written as:
\[
\text{PPL}_\theta(\mathbf{x}) = P_\theta(\mathbf{x})^{-\frac{1}{n}}.
\]

\subsection*{Interpretation}
\begin{itemize}
    \item \textbf{Lower PPL} indicates that the model assigns higher probability to the actual observed sequence $\rightarrow$ \textit{better prediction}.
    \item \textbf{Higher PPL} means the model is more uncertain about the next token $\rightarrow$ \textit{poorer performance}.
\end{itemize}

\section{Negative Log-Likelihood (NLL)}
\label{sec:nll}

Let $x$ be the true token or label, and let $p_\theta(x)$ represent the predicted probability distribution over the token space, parameterized by $\theta$. The log-likelihood of observing $x$ under this distribution is given by\footnote{\url{https://www.statlect.com/glossary/log-likelihood}}:
\[
L(\theta|x) = \log [p_\theta(x)]
\]

A higher log-likelihood indicates that the model assigns a higher probability to the correct token, implying greater confidence in its prediction.

Since probabilities lie between 0 and 1, their logarithms are typically negative. To convert this into a more interpretable loss function, we take the negative log-likelihood (NLL), defined as:
\[
NL(\theta|x) = -\log p_\theta(x)
\]

A lower NLL value implies better model performance, as it means the predicted distribution assigns higher probability to the correct label. NLL can also be interpreted as a penalty term: the less confident the model is in its correct prediction, the higher the penalty it incurs.

\subsection{NLL Behavior in Sequences}
\label{sec:nll-sequences}

In the context of text generation, we are often interested in modeling sequences of tokens. Suppose the true token sequence\footnote{\url{https://datajello.com/cross-entropy-and-negative-log-likelihood/?utm_source=chatgpt.com}} is given by $x = (x_1, x_2, \ldots, x_T)$. The NLL over the entire sequence becomes:
\[
NL(\theta|x) = - \sum_{t=1}^{T} \log p_\theta(x_t \mid x_{<t})
\]

Here, each token $x_t$ is predicted based on all preceding tokens $x_{<t}$. The term $x_{<t}$ denotes the subsequence of tokens before position $t$, and is drawn from a valid subset of the sequence space $\mathcal{R}$. This autoregressive formulation is fundamental to many language models, where prediction quality is evaluated sequentially.

\section{Autoregressive model (AR)}
\label{sec:ar}
Natural language processing (NLP) relies heavily on autoregressive (AR) models, especially when it comes to sequence generation tasks.\cite{bhandari2023survey} These models successfully capture the conditional dependencies inherent in language by generating text by predicting each token based on the order of preceding tokens. AR models can generate logical and contextually relevant text outputs thanks to this sequential prediction technique.\footnote{\url{https://medium.com/@zaiinn440/autoregressive-models-for-natural-language-processing-b95e5f933e1f}}

From Recurrent Neural Networks (RNNs) to more complex structures like Long Short-Term Memory (LSTM) networks and Gated Recurrent Units (GRUs), the architecture ~\ref{ar:architecture} of AR models has changed over time. These advancements have improved the models' capacity to represent long-range dependencies in text and have tackled issues like the vanishing gradient problem. \ref{mdlm} Because of their attention mechanisms and parallel processing capabilities, Transformer-based models—such as the Generative Pre-trained Transformer (GPT) series—have demonstrated notable success in a variety of NLP tasks in recent years.\cite{ramachandran2017fast} 

AR models have been effectively applied in diverse NLP applications, including machine translation, text summarization, and dialogue systems. Their ability to model the probability distribution of sequences makes them particularly well-suited for tasks that require the generation of fluent and contextually appropriate text.

In the subsequent sections, we delve deeper into the mechanics of autoregressive (AR) models, examining their underlying architecture, benefits, limitations, and performance in comparison to alternative modeling approaches. \ref{mdlm}  \ref{ar:architecture}

Autoregressive models factorize the joint probability of a sequence into a product of conditional probabilities, allowing each token to be generated based on all preceding tokens. This is mathematically formulated as: \cite{ramachandran2017fast} 

\begin{align}
P(x_1, x_2, x_3, \ldots, x_n) &= P(x_1) \cdot P(x_2 \mid x_1) \cdot P(x_3 \mid x_1, x_2) \nonumber \\
&\quad \cdots P(x_n \mid x_1, x_2, x_3, \ldots, x_{n-1})
\end{align}

To aid understanding, consider the simple sentence: \textit{"The cat sleeps."} The probability of generating this sentence using an AR model can be expressed as:

\begin{align}
P(\text{"The cat sleeps"}) = P(\text{"The"}) \cdot P(\text{"cat"} \mid \text{"The"}) \cdot P(\text{"sleeps"} \mid \text{"The cat"})
\label{eq:ar-example}
\end{align}

\begin{figure}[h]
    \centering
    \includegraphics[width=0.9\textwidth]{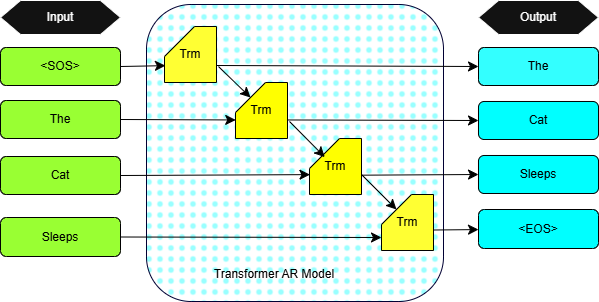}
    \caption{Illustration of the token-by-token generation process in an autoregressive model. This Trm represent the Transformer block.}
    \label{fig:ar-process}
\end{figure}

For a clearer interpretation of Equation~\ref{eq:ar-example}, a visual representation of the AR model’s token generation process is provided in Figure~\ref{fig:ar-process}.

As shown in Figure~\ref{fig:ar-process}, the AR model begins generation with a special \texttt{<SOS>} (start-of-sentence) token. Given this token as input, the model predicts the first word $x_1$. The process continues sequentially, where each subsequent word is generated based on the previous context. The generation process concludes with the \texttt{<EOS>} (end-of-sentence) token, signaling the end of the sequence. This directional and sequential nature of AR models is fundamental to their architecture and influences both their strengths and limitations in language generation tasks. \footnote{\url{https://huggingface.co/docs/transformers/v4.32.1/llm_tutorial?utm_source=chatgpt.com}}

\section{Discrete Denoising Diffusion Probabilistic Models (D3PM)}
\label{sec:d3pm}

D3PM (Discrete Denoising Diffusion Probabilistic Models) represents a class of diffusion models tailored for discrete data types such as text, categorical variables, and token sequences, rather than continuous data like images. In general, diffusion models operate by introducing structured noise into the data over several steps and then learning to reverse this process—i.e., denoising—during generation.\cite{lou2023discrete}  

The training process of a diffusion model begins with tokenized input data. Over successive time steps, increasing levels of noise are added to the data in a controlled manner. The model is then trained to learn the reverse process: gradually denoising the corrupted input to reconstruct the original data. Notably, this denoising process is executed in parallel, making it inherently different from the sequential approach seen in autoregressive models.

The primary objective during training is to minimize the loss function that quantifies the difference between the original data and the model's reconstruction at each denoising step. Through this optimization process, the model progressively improves its generative capability by learning a high-fidelity reverse diffusion path.

We can demonstrate, from a statistical standpoint, that the training objective in diffusion models involves minimizing a loss function derived from the \textbf{Variational Lower Bound (VLB)}~\cite{ho2020denoising}. This objective enables the model to learn the denoising process in a step-by-step manner by approximating the reverse transitions from noisy to clean data.

The variational lower bound loss, often referred to as the negative Evidence Lower Bound (NELBO), is expressed as:~\cite{sahoo2024simple} 

\begin{align}
\mathcal{L}_{\text{vb}} = 
\underbrace{ \mathbb{E}_{q}\left[ D_{\text{KL}}\left(q(x_T \mid x_0) \,\|\, p_\theta(x_T)\right) \right] }_{L_T}
-
\underbrace{ \log p_\theta(x_0 \mid x_1) }_{L_0}
+ 
\nonumber \\
\sum_{t=2}^{T} 
\underbrace{ D_{\text{KL}}\left(q(x_{t-1} \mid x_t, x_0) \,\|\, p_\theta(x_{t-1} \mid x_t)\right) }_{L_{t-1}}
\label{eq:d3pm-error}
\end{align}

where $\mathcal{L}_{\text{vb}}$ denotes the variational loss, and $D_{\text{KL}}[\cdot]$ represents the Kullback–Leibler divergence, a measure of dissimilarity between two probability distributions.

Equation~\ref{eq:d3pm-error} decomposes the total training loss into three interpretable components:

\begin{itemize}
    \item $L_T$: This term measures the KL divergence between the noisiest state $q(x_T \mid x_0)$ and the prior $p_\theta(x_T)$, capturing how well the prior models the final corrupted distribution.
    
    \item $L_0$: This is the negative log-likelihood of reconstructing the original data $x_0$ given the first denoised state $x_1$. It quantifies the reconstruction loss at the initial denoising step and represents the final output loss of the reverse process.

    \item $\sum_{t=2}^{T} L_{t-1}$: These intermediate KL divergence terms assess how closely the learned reverse transition $p(x_{t-1} \mid x_t)$ approximates the true reverse posterior $q(x_{t-1} \mid x_t, x_0)$. They span from $t = 2$ to $T$, covering the core denoising steps in the reverse process.
\end{itemize}

Collectively, this formulation ensures that the model learns to generate data by progressively reversing the corruption process introduced during the forward diffusion. The accuracy of reconstruction, particularly after masking, is reflected in $L_T$, while $L_0$ directly affects output fidelity. The intermediate terms govern the smoothness and consistency of the denoising trajectory throughout the diffusion steps.

Minimizing this loss corresponds to teaching the model to reverse the forward noising process — effectively learning how to denoise and reconstruct the original data.

Diffusion models are characterized by two fundamental processes: the \textbf{forward process} and the \textbf{backward process}. These two stages form the core mechanism of how diffusion models are trained and used for generation.

In the \emph{forward process}, noise is incrementally added to the input data across a series of timesteps, gradually transforming the data into a noise-like distribution. This process is carefully designed to be mathematically tractable and often ends with a uniform or Gaussian distribution, depending on the domain (discrete or continuous).

The \emph{backward process}, on the other hand, involves learning to reverse this corruption by progressively denoising the data to recover the original structure. One of the prominent features of diffusion models is their ability to generate data in a non-sequential manner, allowing them to sample tokens from left-to-right, right-to-left, or even bidirectionally. This flexibility stands in contrast to autoregressive models, which strictly generate text in a left-to-right fashion due to their dependency on previously generated tokens.

In the following section, we will examine the forward process in more detail.

\subsection{Forward Process}
\label{d3pm:forward}

\begin{figure}[h]
    \centering
    \includegraphics[width=0.9\textwidth]{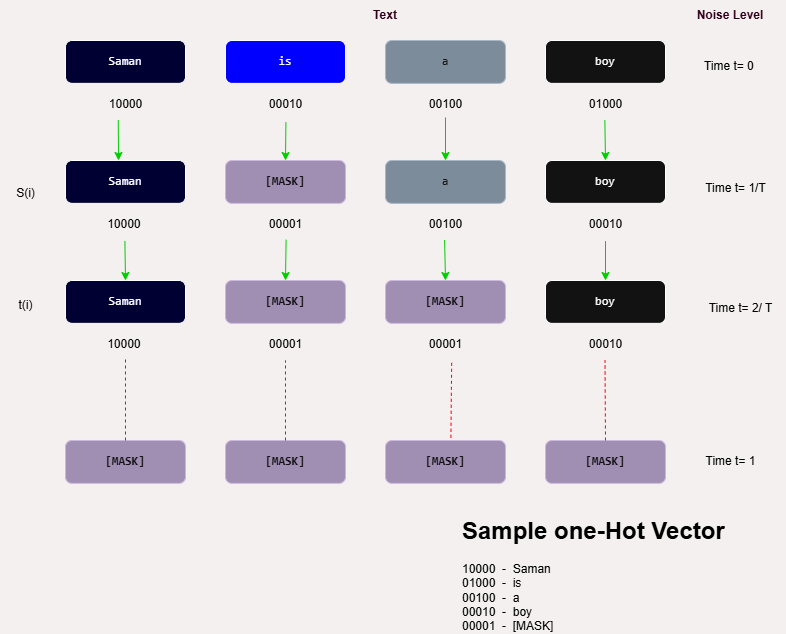}
    \caption{Illustration of the D3PM forward process.}
    \label{fig:d3pm-process}
\end{figure}

In this section, we delve into the technical and mathematical foundations of the diffusion model, moving beyond the high-level overview. To facilitate understanding, we introduce a simple example: the sentence \textit{"Saman is a boy"}, as illustrated in Figure~\ref{fig:d3pm-process}.

Formally, we assume that our data consists of tokens from a finite discrete state space with $m$ possible states, represented by integers $0, 1, \ldots, m-1$ and their corresponding one-hot vectors \footnote{A one-hot vector is a binary vector in which only one element is set to 1 (hot) and all others are 0, typically used to represent categorical data or discrete tokens.} $e_0, e_1, \ldots, e_{m-1}$. In the case of the example, the integers would be $0, 1, 2, 3$, and the corresponding one-hot vectors would be $e_0, e_1, e_2, e_3$.  

To accommodate the masking process, we augment the discrete state space by adding an additional mask state, denoted by the index $m$. During the forward process, each token transitions independently to the mask state at a random time step. This process, applied independently across tokens (e.g., each word), progressively corrupts the original sequence into a sequence dominated by mask tokens. By learning to reverse this corruption, the model can reconstruct coherent discrete data, enabling efficient generation.

We begin by considering the case of a single token before generalizing to multiple dimensions. In discrete diffusion, we simulate a process where information, such as a text token, gradually degrades into noise over $T$ discrete steps. To control and learn this process effectively, the total time interval $[0, 1]$ is divided into $T$ equal segments. The start and end times of the $i$-th interval are defined as:
\[
s(i) = \frac{i - 1}{T}, \quad t(i) = \frac{i}{T}.
\]
Here $s$ denotes the starting timestep, and $t$ represents the subsequent timestep that follows $s$.
According to Austin et al\cite{austin2021structured}, the transition matrix at each step, of size $(m+1) \times (m+1)$, is given by:
\begin{equation}
Q_i = (1 - \beta_i) I + \beta_i \mathbf{1} e_m^\top,
\label{eq:forward_diffusion-q}
\end{equation}

where $\mathbf{1}$ is an all-one column vector of size $m+1$, and $e_m$ is a one-hot vector with the $m$-th element set to 1, representing the mask state. And $Q_i$ could be s or t step in the moment.

As an illustrative example, assume $\beta_i = 0.3$. The corresponding $Q_i$ matrix becomes:
\[
Q_i =
\begin{bmatrix}
0.7 & 0   & 0   & 0   & 0.3 \\
0   & 0.7 & 0   & 0   & 0.3 \\
0   & 0   & 0.7 & 0   & 0.3 \\
0   & 0   & 0   & 0.7 & 0.3 \\
0   & 0   & 0   & 0   & 1.0 \\
\end{bmatrix}
\]

In this matrix, if we consider row 0, the token "Saman" (One-hot vec: [1,0,0,0,0] and $Vec.Q_i$) remains itself with a probability of 0.7, while it transitions to the mask state with a probability of 0.3. Therefore, the next token is most likely to remain "Saman," but may become masked with some probability.

Each entry $[Q_i]_{jk}$ denotes the probability of transition from state $j$ to state $k$:
\[
[Q_i]_{jk} = q(x_{t(i)} = k \mid x_{s(i)} = j).
\]
This means that, with probability $1 - \beta_i$, $x_{t(i)} = x_{s(i)}$, and with probability $\beta_i$, the token transitions to the mask state.

Now, let us consider the behavior of the diffusion process between two arbitrary time steps $s$ and $t$:

\begin{equation}
q(x_{t} \mid x_{s}) = \text{Cat}(x_{t}; \bar{Q}_{t|s}^\top x_{s}) = x_s^\top \bar{Q}_{t|s} x_{t},
\label{eq:forward_diffusion-qxts}
\end{equation}
where the distribution is conditioned on $x_s$, making the marginal naturally expressed using $x_s^\top$. Here, $\bar{Q}_{t|s}$ denotes the transition matrix from time step $s$ to $t$, which represents the cumulative effect of corruption between these two time steps. The transpose of the transition matrix is used to define the categorical distribution over the target token $x_t$, ensuring compatibility with the form of the categorical distribution parameterized by $p$.

Then, we assign Equation~\ref{eq:forward_diffusion-q} to Equation~\ref{eq:forward_diffusion-qxts}, which describes the categorical transition probability from state $x_s$ to $x_t$:

\[
q(x_t \mid x_s) = \text{Cat}\left(x_t; \left[(1 - \beta_{t|s}) I + \beta_{t|s} \mathbf{1} e_m^\top \right]^\top x_s \right)
\]

This can be simplified using matrix-vector multiplication:

\[
= \text{Cat}(x_t; \, (1 - \beta_{t|s}) x_s + \beta_{t|s} e_m)
\]

This implies the following behavior for $x_t$:

\begin{equation}
x_t = 
\begin{cases}
(1 - \beta_{t|s}) , & \textit{if the original token is retained} \\
\beta_{t|s},       & \textit{if the token is masked} \\
0,                     & \textit{otherwise}
\end{cases}
\end{equation}

The above case-based formulation simplifies the behavior of the forward diffusion process. In the context of this research, it is important to note that as diffusion progresses, the probability term $\beta_{t|s} e_m$ approaches 1, since at the end of the forward process, all tokens are expected to be fully masked.

The key insight here lies in the relative magnitude of the two probability components: $(1 - \beta_{t|s}) x_s$ and $\beta_{t|s} e_m$. Whether the former is greater than the latter determines the likelihood of a token being preserved versus being masked. In essence, this trade-off governs the future behavior of tokens during diffusion, influencing the model's capacity to reconstruct or predict them accurately in the reverse process.

Now, let us turn our attention to how the model estimates unknown tokens from the initial clean input. Given the forward transition matrices, the marginal distribution at any time step $t(i)$, conditioned on the original token $x_0$, is expressed as:

\begin{equation}
q(x_{t(i)} \mid x_0) = \text{Cat}(x_{t(i)}; \bar{Q}_t^\top x_0) = x_0^\top \bar{Q}_t x_{t(i)}, \quad \text{where} \quad \bar{Q}_t = Q_1 Q_2 \cdots Q_t,
\label{eq:forward_diffusion}
\end{equation}

where $\text{Cat}(x, p)$ denotes the categorical distribution over the values of $x$, with $p$ representing the probability vector corresponding to each category. The matrix $\bar{Q}_t$ is the product of all transition matrices from step 1 to $t$, capturing the cumulative noise applied during the forward process. This expression enables the computation of the corrupted distribution at any given time step based on the original clean input.

\subsection{Backward Process}
\label{d3pm:backward}

The backward process constitutes the learnable component of the diffusion model. Its primary objective is to reverse the corruption introduced during the forward process by learning to denoise the data in a step-by-step manner. During training, the model is optimized to reconstruct the original input from its noisy counterparts observed at various time steps. This optimization typically involves minimizing a loss function, such as the expected squared error between the predicted and true noise, thereby enabling the model to adjust its parameters effectively.

\begin{figure}[h]
    \centering
    \includegraphics[width=0.9\textwidth]{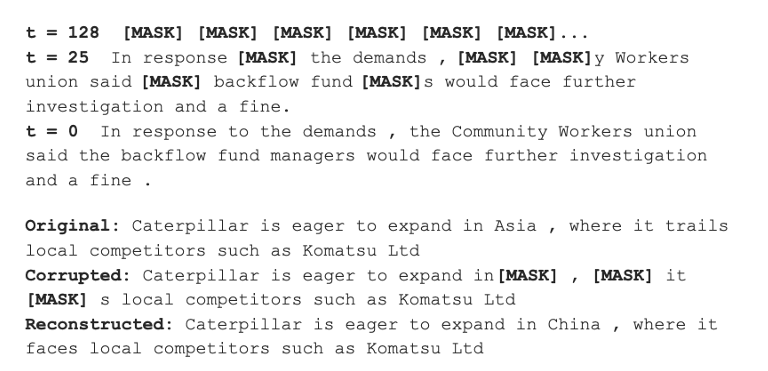}
    \caption{Using a trained D3PM absorbing model for LM1B to (top) generate new sentences and (bottom) reconstruct corrupted examples.}
    \label{fig:d3pm-process-backward}
    \url{https://arxiv.org/pdf/2107.03006}
\end{figure}

To better understand the backward process, let us consider how results can be derived given the original data $x_0$. Owing to the analytical properties of the forward process, several quantities of interest can be computed in closed form. One such important quantity in diffusion models is the time-reversal distribution of the forward process conditioned on the original input, denoted as $q(x_s \mid x_t, x_0)$ for $s < t$. 

Due to the Markov property of the diffusion process, we have:
\[
q(x_{s(i)} \mid x_{t(i)}) = q(x_{s(i)} \mid x_{t(i)}, x_0),
\]
which reflects the fact that the future state depends only on the current state, not on the full history.

Using Bayes' rule and the chain rule,~\cite{sahoo2024simple} the time-reversed posterior can be expressed as:

\begin{align}
q(x_{s(i)} \mid x_{t(i)}, x_0) = \frac{q(x_{t(i)} \mid x_{s(i)}, x_0) \cdot q(x_{s(i)} \mid x_0)}{q(x_{t(i)} \mid x_0)}
\nonumber\\
= \text{Cat}\left( x_{s(i)}; \, p =  \frac{x_{t(i)} \bar{Q}_{t|s}^\top \odot x_0 \bar{Q}_{s}}{x_0 \bar{Q}_{t} x_{t(i)}^\top} \right),
\label{eq:backward_diffusion2}
\end{align}

$ \text{Where }\bar{Q}_{t|s}\triangleq \bar{Q}_t^{-1} \bar{Q}_s$

where $\odot$ denotes element-wise (Hadamard) multiplication, and $Q_{t|s}$ is the transition matrix from time $s$ to $t$. In this formulation, $x_{s(i)}$ remains a categorical variable, and $p$ represents the probability vector over all possible categories.

An equivalent and expanded form of this expression can be written as:

\[
q(x_s \mid x_t, x_0) = \text{Cat}\left( x_s; \, p = \frac{\left[ (1 - \beta_{t|s}) I + \beta_{t|s} \mathbf{1} e_m^\top \right] x_t \odot \left[ \beta_s e_m + (1 - \beta_s) x_0 \right]}{x_t^\top \left[ \beta_s e_m + (1 - \beta_s) x_0 \right]} \right),
\]

where $\beta_s$ and $\beta_{t|s}$ (in here $\beta_{t|s}=\frac{\beta_{t}}{\beta_{s}}$ ) are noise parameters for the respective timesteps, $\mathbf{1}$ is an all-ones vector, $e_m$ is the one-hot mask token, and $I$ is the identity matrix. This formulation captures the probabilistic mechanism used to reverse the forward corruption process during inference.

The process described above is illustrated in Figure~\ref{fig:d3pm-process-backward}, which demonstrates both sentence generation and the reconstruction of corrupted sequences using a trained D3PM model.

The model is designed to approximate the original data $x_0$ using a denoising function parameterized by $\theta$, defined as:

\[
x_\theta(x_t, t): \mathcal{V} \times [0, 1] \rightarrow \Delta^K,
\]

where $\mathcal{V}$ denotes the input space, $[0,1]$ represents the diffusion time interval, and $\Delta^K$ is the $K$-dimensional probability simplex corresponding to a categorical distribution over $K$ classes.

This approximation is then plugged into the posterior distribution to estimate the denoised state:

\[
p_\theta(x_s \mid x_t) = q(x_s \mid x_t, x_0 = x_\theta(x_t, t)),
\]

which forms the foundation of the denoising objective—learning to infer the clean data $x_0$ from its corrupted version $x_t$ through an approximate reverse process.

Consequently, the posterior distribution over the discrete categorical variable $x_s$ can be expressed as:

\[
q(x_s \mid x_t, x_0 = x_\theta(x_t, t)) 
= \text{Cat}\left( x_{s(i)}; \, p = \frac{x_{t(i)} Q_{t|s}^\top \odot x_\theta(x_t, t) \bar{Q}_s}{x_\theta(x_t, t) \bar{Q}_t x_{t(i)}^\top} \right),
\]

where $\odot$ denotes element-wise multiplication. Here, $x_\theta(x_t, t)$ serves as a key learnable approximation of the true data distribution. It is this parameter that is optimized during training, enabling the model to denoise corrupted observations and reconstruct coherent outputs in the backward diffusion process.


 

\chapter{Methodology}
\label{cha:method}

The primary objective of this chapter is to outline the methodological framework used in this research. This includes the experimental design, model selection, data preprocessing pipeline, and evaluation strategy. The methodology aims to provide a transparent, reproducible structure that supports the comparative analysis between Autoregressive (AR) and Discrete Diffusion (D3PM) models for language generation tasks.

Specifically, this chapter is organized into the following sections:
\begin{itemize}
    \item \textbf{Dataset Description} – covering the source, structure, and suitability of the dataset used.
    \item \textbf{Tokenization and Preprocessing} – explaining how the raw text data was transformed into a tokenized format suitable for training.
    \item \textbf{Model Architectures} – describing both AR and D3PM models, their configurations, and the training strategies employed.
    \item \textbf{Evaluation Metrics and Procedure} – detailing how model performance was measured, including the choice of statistical metrics (e.g., BPT, NLL, PPL) and how results were collected and compared.
\end{itemize}

Although certain preplanned experiments could not be completed due to time constraints, these limitations are addressed in the \textit{Future Work} section (\autoref{sec:discussion-Future Work}). Nevertheless, the methodology presented here establishes a consistent and rigorous approach for evaluating and comparing the two model paradigms under fair and controlled conditions.

\section{Dataset}
\label{sec:Method-data}

To compare the effectiveness of D3PM and autoregressive (AR) models in language generation, we evaluate their performance using key metrics such as Bits Per Token (BPT)~\ref{sec:bpt}, Perplexity (PPL)~\ref{sec:ppl}, and Generation Speed. To ensure a fair comparison, all models are trained and tested on the same dataset under identical experimental conditions.

In this study, the WikiText-103 dataset~\cite{merity2016pointer} is selected for evaluation. WikiText-103 is a widely used benchmark for language modeling, known for its rich vocabulary, long-range dependencies, and real-world complexity. The dataset comprises 4,292,663 sentences in the training set, 9,070 in the validation set, and 10,702 in the test set. All experiments in this research are conducted solely on WikiText-103 to ensure consistency and control in performance evaluation.

The model architecture used in this study contains approximately 124 million trainable parameters, making it comparable in scale to mid-sized transformer-based language models.

\begin{figure}[h]
    \centering
    \includegraphics[width=0.9\textwidth]{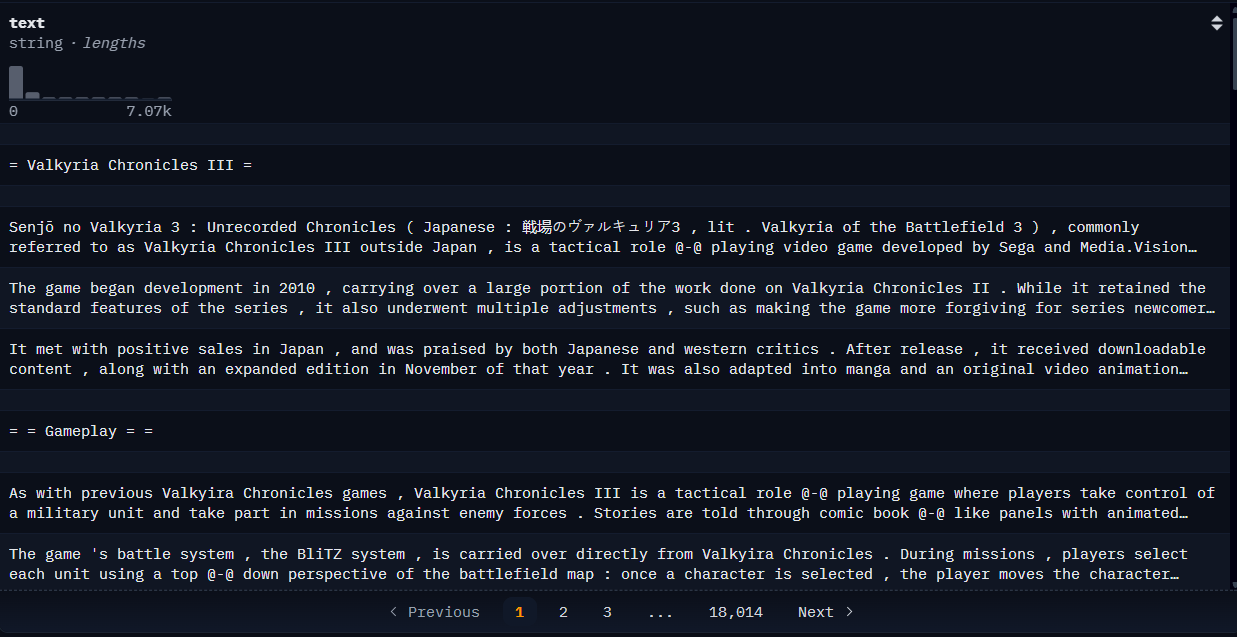}
    \caption{Sample spectrum from the WikiText-103 dataset.}
    \label{fig:wikitext103}
\end{figure}

Compared to the preprocessed Penn Treebank (PTB), WikiText-2 is over 2 times larger, while WikiText-103 is more than 110 times larger. Moreover, unlike PTB—which removes case information, punctuation, and numbers—the WikiText datasets preserve these features, offering a more realistic and linguistically rich corpus. Composed of full Wikipedia articles, WikiText-103 is particularly suitable for evaluating models capable of capturing long-range dependencies in text. \footnote{\url{https://huggingface.co/datasets/Salesforce/wikitext}}

Each split of the dataset is available in two formats:

\begin{itemize}
    \item \textbf{Raw}: Contains raw tokens and is commonly used for character-level tasks.
    \item \textbf{Non-raw}: Contains only vocabulary tokens suitable for word-level tasks, where out-of-vocabulary items are replaced with a designated token.
\end{itemize}

In this study, we utilize the \textbf{Non-raw} version of WikiText-103, as our focus is on word-level language modeling. One of the notable challenges with this dataset is the frequent appearance of symbols such as \texttt{=}, formula-like structures, and special tokens. As illustrated in Figure~\ref{fig:wikitext103}, these artifacts often lead to generated outputs that include irrelevant tokens like “=”, “formula,” or isolated special characters—especially in diffusion-based generation. This behavior reflects the statistical frequency of such patterns in the training data.

Overall, WikiText-103 serves as a suitable and challenging benchmark for evaluating generative models on realistic and diverse natural language input.

\section{Data Pre-Processing}
\label{sec:Method-token}

\subsection{Definition and Importance of Tokenization}
\label{sec:tokenization-definition}

Data Preprocess AKA Tokenization is the process of segmenting a continuous stream of text into discrete units called tokens. These tokens serve as the fundamental building blocks for subsequent natural language processing (NLP) tasks, including morphological analysis, word-class tagging, and syntactic parsing. As described by Habert et al.\cite{habert1996symbolic},  

 "Tokenization, that is, the identification of each 'atomic' unit, represents the very first operation to be performed in document processing; nevertheless, it is often overlooked because of its supposed basic nature."  ~\cite{habert1998towards}.  

Despite its apparent simplicity, tokenization plays a critical role in determining the effectiveness of downstream NLP tasks. Since most NLP pipelines operate at the sentence level, an essential aspect of tokenization is also the identification of sentence boundaries alongside token boundaries~\cite{grefenstette1999tokenization}.  

\subsection{Text to Tokens}
\label{sec:tokenization-application}

\begin{figure}[h]
    \centering
    \includegraphics[width=0.9\textwidth]{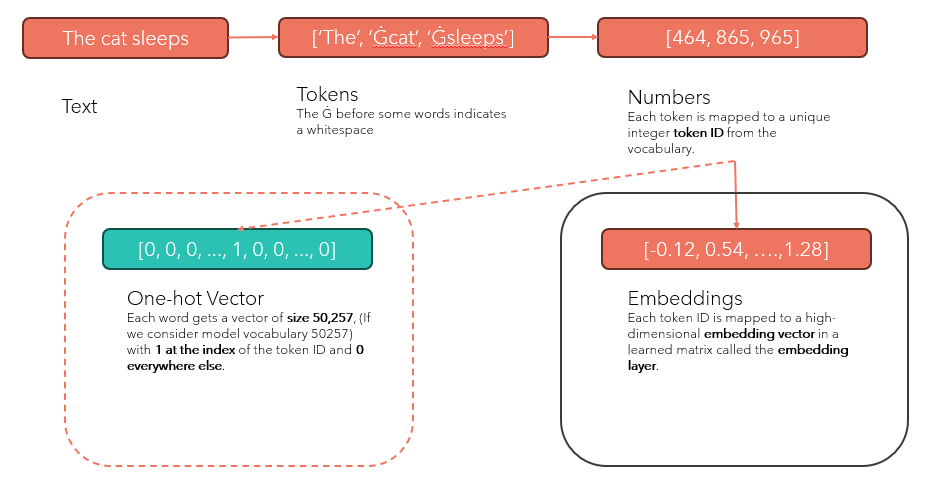}
    \caption{Illustration of the tokenization process used in this project.}
    \label{fig:tokenization-process}
\end{figure}

Tokenization is a crucial preprocessing step in this project, enabling the transformation of raw text into a structured format suitable for neural network-based language models. The process can be broken down into the following stages, as illustrated in Figure~\ref{fig:tokenization-process}.

\begin{itemize}
    \item \textbf{Raw Text Input:} The original sentence, such as "The cat sleeps," is provided as input.
    \item \textbf{Subword Tokenization:} The sentence is divided into smaller units (subwords) rather than whole words, which helps handle rare words and improves model generalization. For example, the sentence is tokenized into ['The', 'Ġcat', 'Ġsleeps'].
    \item \textbf{Token ID Conversion:} Each token is mapped to a unique numerical identifier from a predefined vocabulary. This step ensures compatibility with deep learning models, where text data must be represented as numbers. In this example, the tokenized words correspond to the indices [464, 865, 965] in the vocabulary.
\end{itemize}

This approach, leveraging subword tokenization techniques like Byte Pair Encoding (BPE) \footnote{These are algorithms that learn the most common subword units from a training corpus. For example, BPE merges frequent character pairs like “t” + “h” → “th”, “th” + “e” → “the”, and so on.} or WordPiece, allows the model to efficiently process diverse linguistic structures while maintaining semantic meaning.

\subsection{Token IDs to Embedding Vectors}
\label{sec:tokenization-Embedding}

This step occurs after the tokenization and numerical ID assignment but before the data is fed into the core models. It’s typically part of the input processing pipeline.

\subsubsection{How Does the Embedding Process Work?}
\label{sec:Embedding-work}

\begin{itemize}
    \item \textbf{Embedding Layer:} This is typically a lookup table (e.g., a matrix of shape $[\mathit{vocab\_size}, \mathit{embedding\_dim}]$)   where each row corresponds to the embedding vector for a specific token ID.
    \item \textbf{Implementation:} In practice, this is done using an embedding layer in frameworks like PyTorch or TensorFlow (e.g., $nn.Embedding$ in PyTorch).
    \item \textbf{Example:} As the mentioned in above figure~\ref{fig:tokenization-process} we convert token ID's to multi- dimentional embedding vector complex. The shape of the matrix depend on above mentioned conditions.
\end{itemize}

\section{Models training and evaluation.}
\label{sec:Method-models}

\begin{figure}[h]
    \centering
    \includegraphics[width=0.9\textwidth]{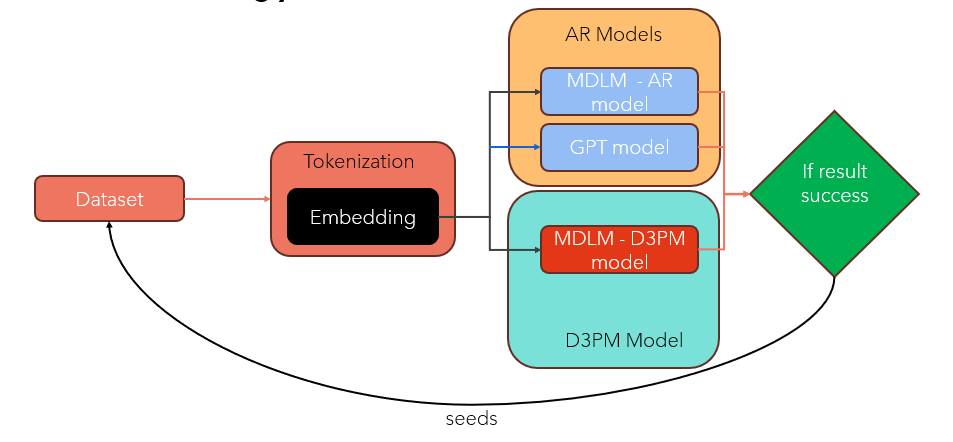}
    \caption{Illustration of the methodology in this project.}
    \label{fig:methodology}
\end{figure}

\subsection{Autoregressive Model Usage}
\label{sec:method-models-ar}

As discussed in Section~\ref{sec:ar}, autoregressive (AR) models generate text token-by-token, conditioning each prediction on previously generated tokens. In this study, for training the AR model, we utilize the tokenized dataset in its embedding vector representation as described earlier.

To enhance the reliability and confidence of our experimental results, we implement an additional validation step by training a GPT-2 model under the same dataset and training conditions. GPT-2, being a strong autoregressive baseline, allows for a meaningful cross-validation of the AR model's performance.

As illustrated in Figure~\ref{fig:methodology}, we systematically compare the performance of the simple autoregressive model with that of GPT-2. This comparative analysis ensures that any observed results are robust and not artifacts of a particular model implementation. By maintaining consistent experimental settings across models, we aim to provide a fair and objective evaluation of the language generation capabilities of autoregressive approaches.

\subsection{D3PM Model Usage}
\label{sec:method-models-d3pm}

In this study, the primary objective when using the D3PM model is to minimize the training loss over categorical data. Specifically, the model aims to reduce the Evidence Lower Bound (ELBO) during training, where a lower ELBO value indicates a more successful learning of the data distribution.

For this purpose, we utilize the \texttt{mdlm} package, developed by Jonathan Ho et al.~\cite{ho2020denoising}, which is designed for training diffusion models on discrete datasets and generating corresponding results. The \texttt{mdlm} framework provides the necessary infrastructure to implement and evaluate discrete diffusion models efficiently.

As illustrated in Figure~\ref{fig:methodology}, the performance of the D3PM model is systematically compared with the autoregressive model results. This comparison allows for an in-depth analysis of the strengths and weaknesses of each approach under consistent experimental conditions.

\subsection{Evaluation and Discussion}
\label{sec:method-results}

In this study, we conduct a detailed evaluation by training each model under different random seed values to observe the variability and robustness of their performance. For both model types, we systematically assess several key metrics: Bits Per Token (BPT)~\ref{bpt:bpc to bpc}, Perplexity (PPL)~\ref{sec:ppl}, Negative Log-Likelihood (NLL)\ref{sec:nll}, generation speed (measured in tokens per second), and the mean and standard deviation of BPT across different runs.

These quantitative results enable us to draw logical conclusions regarding the advantages and limitations of each modeling approach. By comparing these metrics across models, we can identify trends related to efficiency, fluency, and confidence in generation.

Furthermore, based on the collected data, we engage in a detailed discussion to interpret the results critically, highlighting performance trade-offs, model behaviors, and practical implications. Finally, to complement the quantitative evaluation, we propose conducting a human evaluation by analyzing a selection of generated sample texts. This human judgment adds qualitative insight into aspects such as coherence, relevance, and naturalness that are difficult to fully capture with automated metrics.




\chapter{Results}
\label{cha:results}

In this chapter, we present and analyze the experimental results obtained from the training and evaluation of both the Autoregressive (AR) and D3PM models. Before delving into the individual outcomes, it is important to highlight the standardized setup and evaluation protocol used to ensure a fair comparison across all models.

All models were trained on the \textbf{WikiText-103} dataset using a total diffusion time horizon of $T = 1000$. During training, each model was run with a batch size of 4, and training was performed on the same high-performance computing (HPC) infrastructure to ensure consistency in computational resources.

For the evaluation phase, the number of generated tokens was limited to 100{,}000 per model to ensure uniformity in performance comparison. Additionally, a batch size of 4 was consistently used across all evaluations. This constraint allowed for the reliable measurement of model throughput in terms of batch processing speed (i.e., batches per second), enabling a fair comparison of computational efficiency.

To further investigate model robustness and variance, multiple training runs were conducted using different random seed values: 1, 12, 1000, 2000, and 3000. Although the seed values influence model initialization and sampling paths, they do not impact the core architecture or training configuration. Each seed value thus corresponds to an independently trained instance of the same model type, allowing us to assess how sensitive the models are to initialization and to examine consistency in performance across different training runs.

By maintaining these constraints, we ensure that the results reflect model behavior rather than variability introduced by hardware, dataset partitioning, or runtime parameters. The following sections provide detailed comparisons across key performance metrics, including Bits Per Token (BPT)\ref{sec:bpt}, Negative Log-Likelihood (NLL)\ref{sec:nll}, Perplexity (PPL)\ref{sec:ppl}, and generation speed.

\section{Autoregressive Model results}
\label{sec:results-ar}

After training the models with multiple random seeds, the resulting performance metrics were analyzed. Before discussing the outcomes, it is essential to define the interpretation, typical ranges, and limitations associated with the following evaluation metrics:

\begin{itemize}
    \item \textbf{Bits Per Token (BPT)}: A lower BPT value indicates better compression and prediction capabilities of the model. Typically, a BPT value below 3.5 reflects high efficiency in modeling, while values above 4.5 suggest poor compression and weaker predictive performance.

    \item \textbf{Negative Log-Likelihood (NLL)}: This metric assesses the confidence of the model's predictions. An NLL value below 2.5 suggests the model is highly confident and accurate, whereas values above 3.5 may indicate that the model is struggling to learn meaningful representations from the data.

    \item \textbf{Perplexity (PPL)}: Perplexity evaluates the fluency of the model's text generation. A perplexity score below 20 generally corresponds to fluent and coherent text generation. In contrast, scores above 40 may imply inconsistent, repetitive, or unnatural outputs, indicating weaker generalization.

    \item \textbf{Training Speed (Batch/s)}: Here, we define training speed in terms of batches processed per second. A value below 30 batches/s may indicate potential bottlenecks due to large model size, slow data pipelines, or hardware inefficiencies. In contrast, speeds exceeding 100 batches/s suggest a highly optimized training setup suitable for large-scale datasets.
\end{itemize}

\begin{table}[h]
\centering
\caption{Performance Comparison of AR and GPT-2 Models}
\begin{tabular}{|l|r|r|r|r|r|r|}
\hline
\textbf{Model} & \textbf{Seed} & \textbf{BPT} & \textbf{NLL} & \textbf{PPL} & \textbf{Speed (Batch/s)} & \textbf{Mean of BPT} \\
\hline
AR  & 1    & 4.5925 & 3.1833 & 24.126  & 3.1832  & \multirow{2}{*}{4.59765} \\ 
    & 12   & 4.6028 & 3.1904 & 24.2989 & 3.1904  &                          \\
\hline
GPT-2 & 1000 & 4.2546 & 2.9491 & 19.0886 & 6.74  & \multirow{2}{*}{4.2547}  \\
      & 2000 & 4.2548 & 2.9492 & 19.0905 & 6.83  &                          \\
\hline
\end{tabular}
\label{tab:ar-model_comparison}
\end{table}

In above table \ref{tab:model_comparison} our main purpose is extracting Autoregressive model correct specifications and evaluating with GPT2 model. So we take GPT2 model results as a reference for the Autoregressive model.

After running the AR model and GPT-2 model, we visualized the training statistics using two different seeds for each model. The graphs below illustrate the behavior of key training metrics.

\begin{figure}[h]
    \centering
    \includegraphics[width=0.9\textwidth]{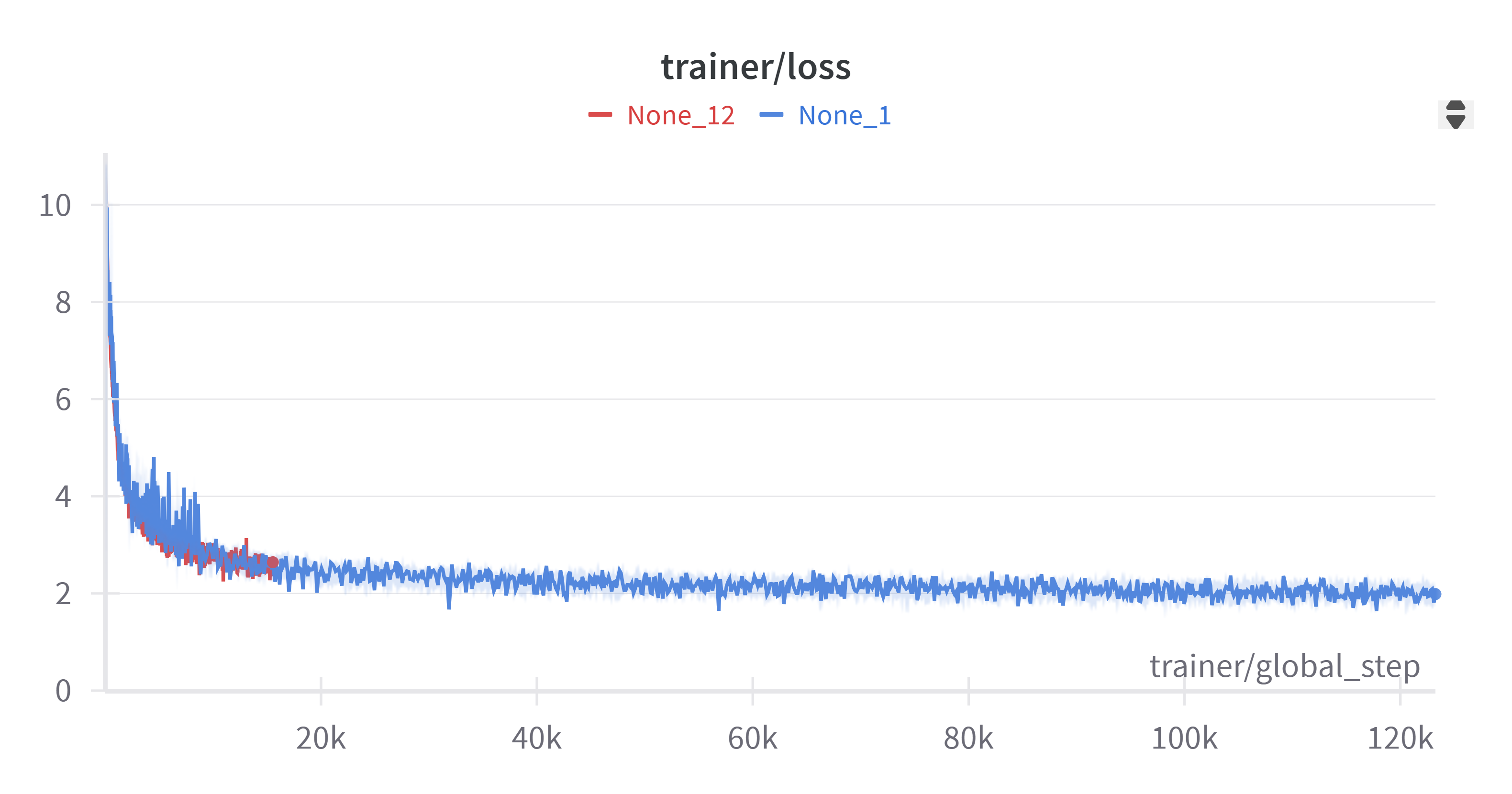}
    \caption{AR model training loss (seeds 1 and 12).}
    \label{fig:ar-train-loss}
\end{figure}

\begin{figure}[h]
    \centering
    \includegraphics[width=0.9\textwidth]{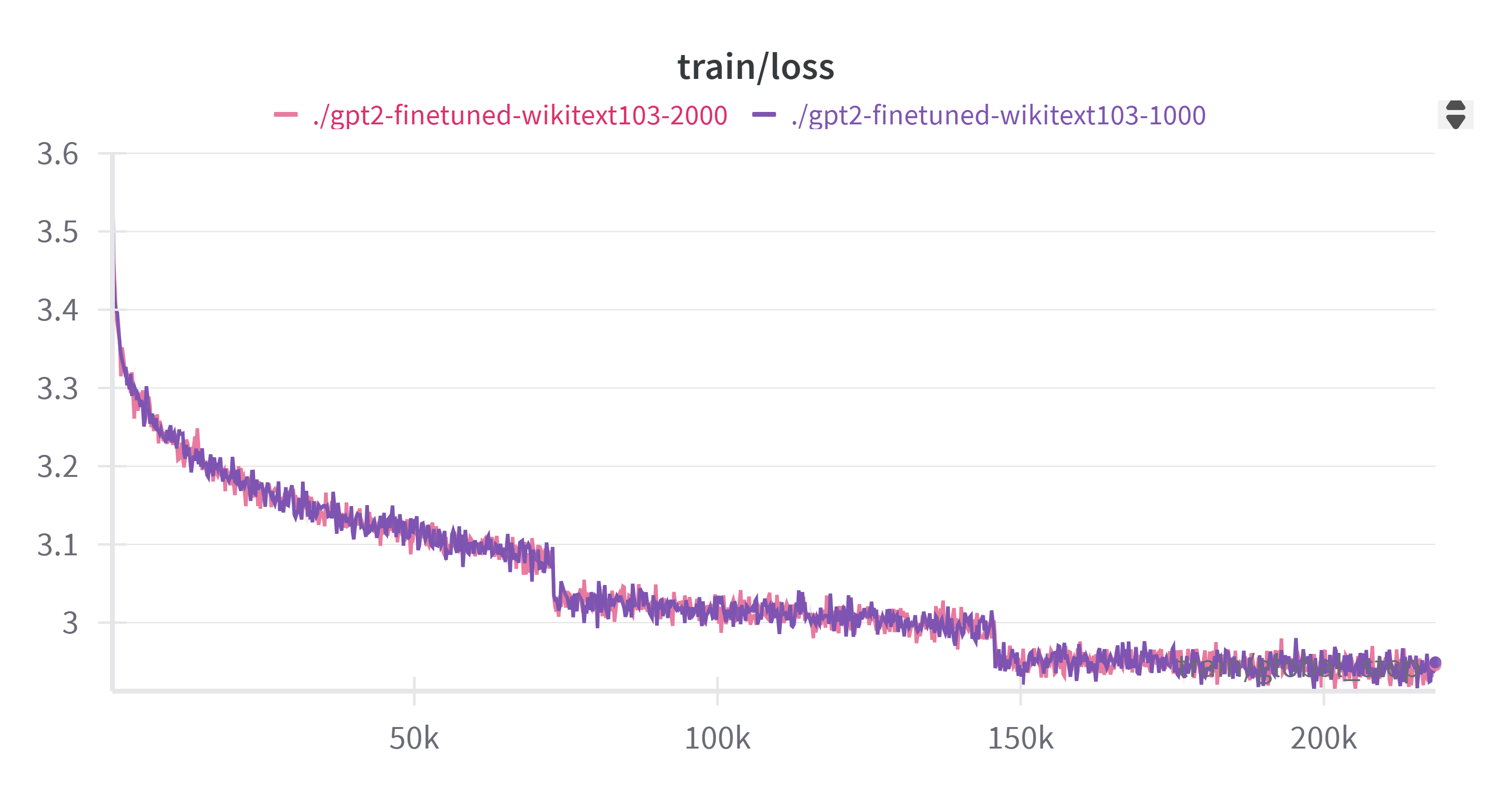}
    \caption{GPT-2 model training loss (seeds 1000 and 2000).}
    \label{fig:gpt-train-loss}
\end{figure}

The training loss graphs represent how well the models are learning over time. The X-axis shows training epochs, and the Y-axis shows the loss value, typically cross-entropy. A lower loss value indicates that the predicted probability distributions are becoming closer to the true token distributions.

\begin{figure}[h]
    \centering
    \includegraphics[width=0.9\textwidth]{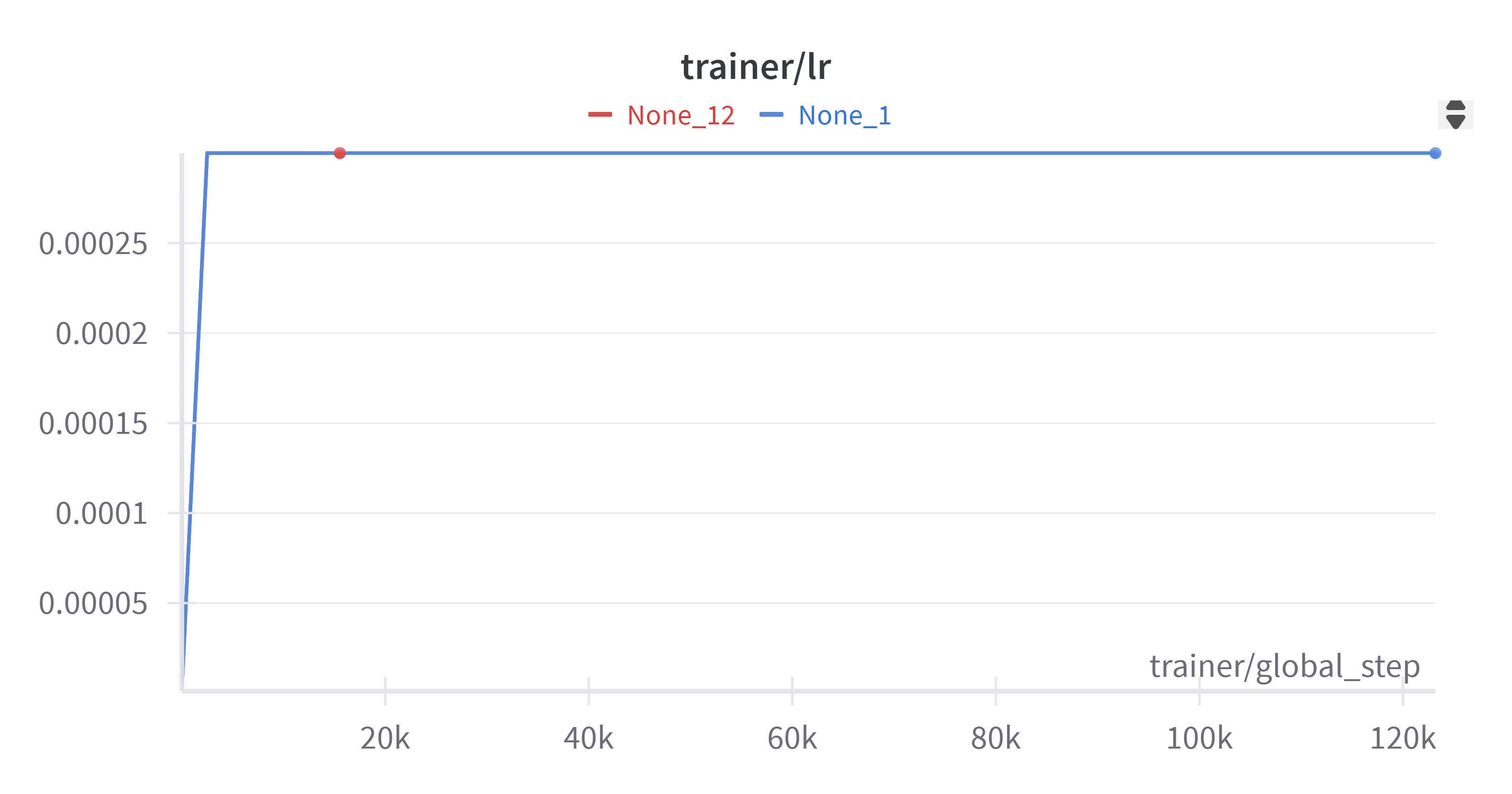}
    \caption{Learning rate schedule for AR model.}
    \label{fig:ar-train-lr}
\end{figure}

\begin{figure}[h]
    \centering
    \includegraphics[width=0.9\textwidth]{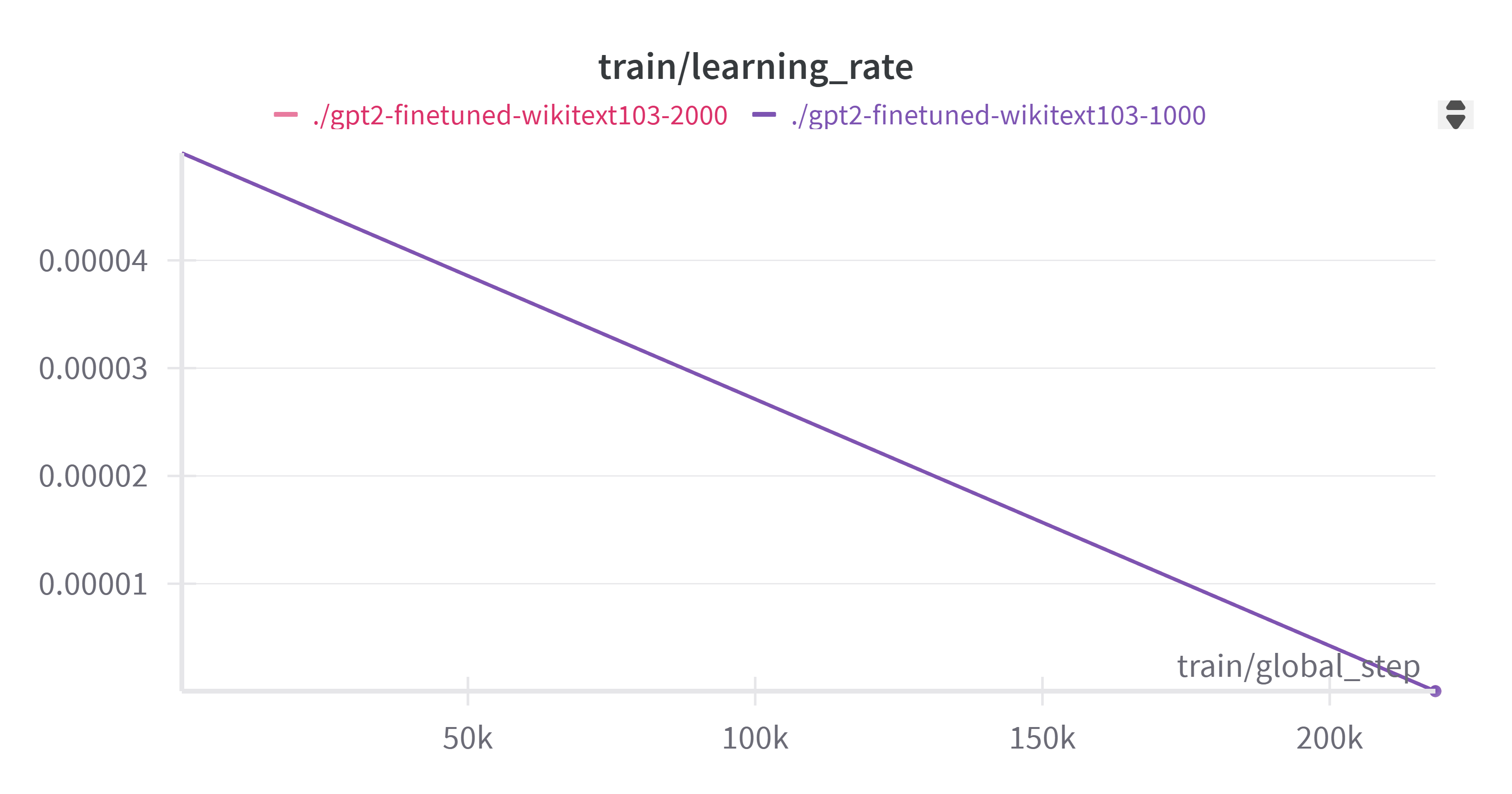}
    \caption{Learning rate schedule for GPT-2 model.}
    \label{fig:gpt-train-lr}
\end{figure}

The learning rate graphs illustrate how the optimizer's learning rate evolves during training. The X-axis represents the global training steps, while the Y-axis shows the learning rate. These schedules often include:
\begin{itemize}
    \item \textbf{Linear warmup:} Learning rate gradually increases at the start.
    \item \textbf{Decay (e.g., cosine, step):} Learning rate decreases as training progresses.
\end{itemize}

\begin{figure}[h]
    \centering
    \includegraphics[width=0.9\textwidth]{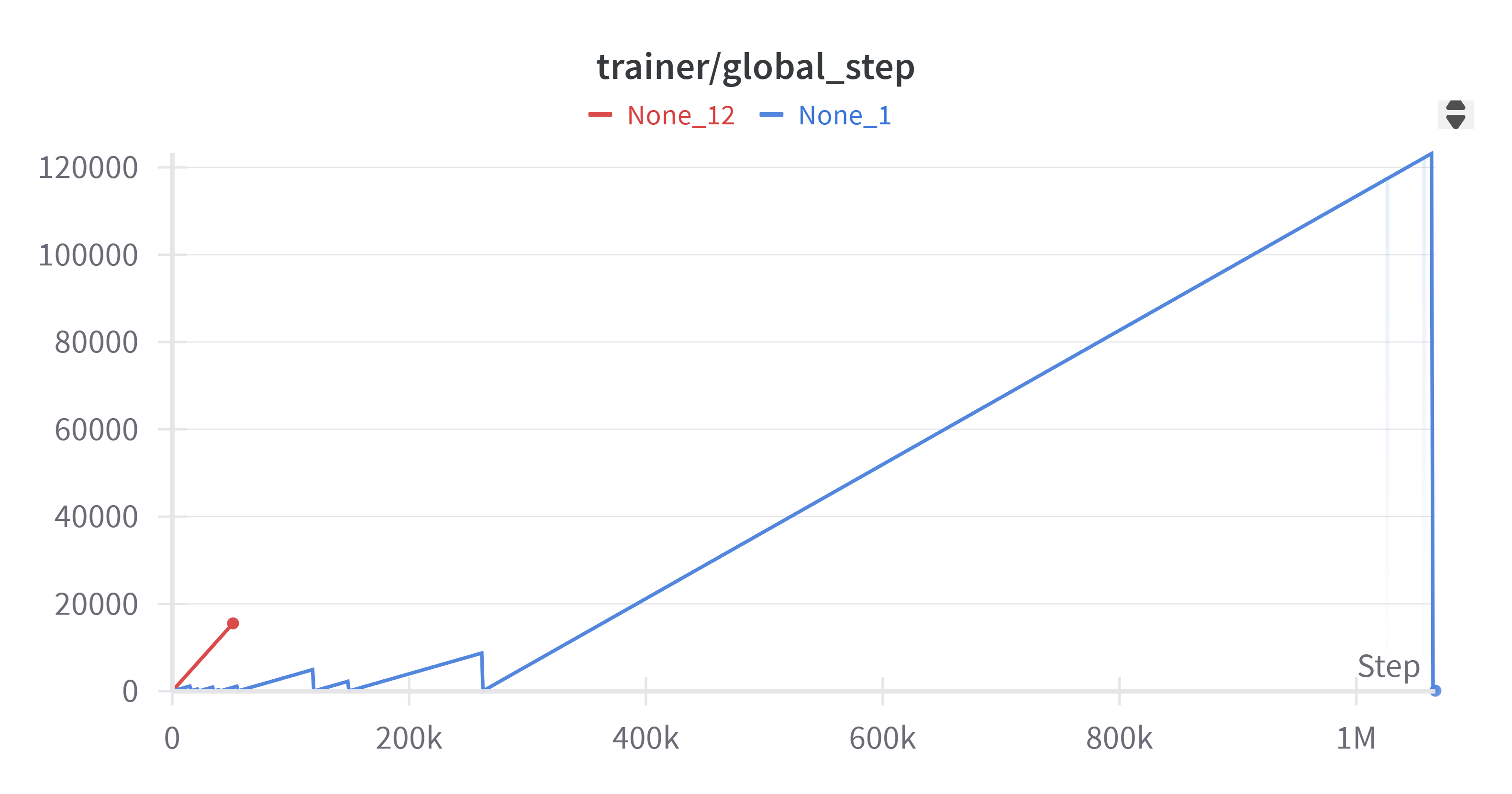}
    \caption{Global step progression in AR model training.}
    \label{fig:ar-train-globalstep}
\end{figure}

\begin{figure}[h]
    \centering
    \includegraphics[width=0.9\textwidth]{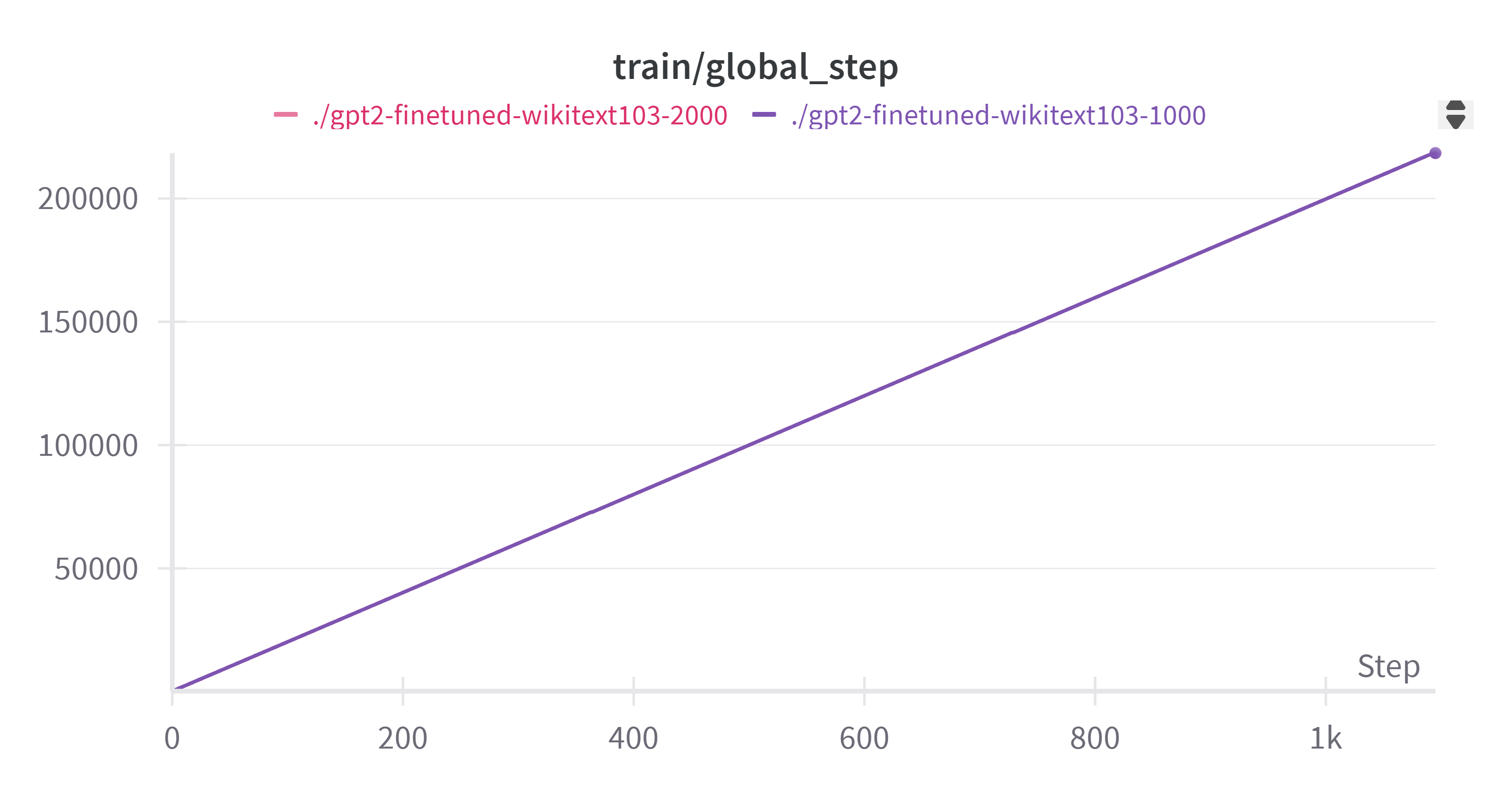}
    \caption{Global step progression in GPT-2 model training.}
    \label{fig:gpt-train-globalstep}
\end{figure}

Figures~\ref{fig:ar-train-globalstep} and~\ref{fig:gpt-train-globalstep} illustrate the global step progression throughout the training of AR and GPT-2 models, respectively. The global step metric represents the cumulative number of optimization steps taken during training, where each step corresponds to processing a single batch and updating model weights via backpropagation.

A notable distinction can be observed between the two plots. In the GPT-2 training (Figure~\ref{fig:gpt-train-globalstep}), the global step progression appears consistent across seeds, with minimal deviation. In contrast, the AR training curve (Figure~\ref{fig:ar-train-globalstep}) exhibits irregularity and overlap in step progression. This discrepancy arises from an earlier test run conducted with the AR model using seed 1 to validate the HPC configuration. The subsequent continuation of training under the same seed led to cumulative steps being visualized as overlapping segments in the graph.

Monitoring global step progression is crucial for diagnosing training stability, determining appropriate learning rate schedules, and identifying the need for gradient clipping. Variations in this metric often reflect changes in system performance, checkpoint resumption, or experimental design.

\section{D3PM Results}
\label{sec:results-d3pm}

In evaluating the D3PM model, we adopted the same evaluation criteria used for the AR and GPT-2 models to ensure a consistent comparison framework \ref{mdlm}. These criteria include Bits Per Token (BPT), Negative Log-Likelihood (NLL), Perplexity (PPL), and generation speed measured in batches per second.

Unlike the AR and GPT-2 models, which had multiple seed-based evaluations and comparative baselines, D3PM lacks a predefined reference point in this experimental setting. Therefore, to establish a robust understanding of its performance, we conducted three separate training runs using different seed values: 1000, 2000, and 3000.

\begin{table}[h!]
\centering
\caption{Evaluation Results for D3PM on WikiText103}
\begin{tabular}{|l|c|c|c|c|c|}
\hline
\textbf{Dataset} & \textbf{Seed} & \textbf{BPT} & \textbf{NLL} & \textbf{PPL} & \textbf{Speed (Batch/s)} \\
\hline
\multirow{3}{*}{WikiText103} 
  & 1000 & 9.0440     & 6.2693     & 528.1280      & 6.2690    \\
  & 2000 & 5.7219     & 3.9661     & 52.7819       & 3.9661    \\
  & 3000 & 9.4063     & 6.5199     & 678.5501      & 6.5199    \\
\hline
\multicolumn{2}{|c|}{\textbf{Mean BPT}} & \textbf{8.0574} & -- & -- & -- \\
\hline
\end{tabular}
\label{tab:d3pm-model_result}
\end{table}

As shown in Table~\ref{tab:d3pm-model_result}, the performance of D3PM varies significantly across different seed values. The model trained with seed 2000 yielded the best results in terms of BPT, NLL, and PPL, suggesting greater stability and convergence in that run. In contrast, models trained with seeds 1000 and 3000 exhibited considerably higher error metrics, pointing to instability or training divergence.

This highlights the importance of conducting multiple seed-based evaluations, especially when dealing with complex generative models like D3PM that lack standardized training configurations. While the mean BPT value across runs is relatively high at 8.0574,the run with seed 2000 yielded the most stable results, with moderate BPT, NLL, and perplexity values. In contrast, the run with seed 3000 diverged significantly, resulting in very high loss and perplexity values. This suggests that the D3PM model is highly sensitive to initialization and may require careful tuning of hyperparameters or improved training stability mechanisms.

\begin{figure}[h]
    \centering
    \includegraphics[width=0.9\textwidth]{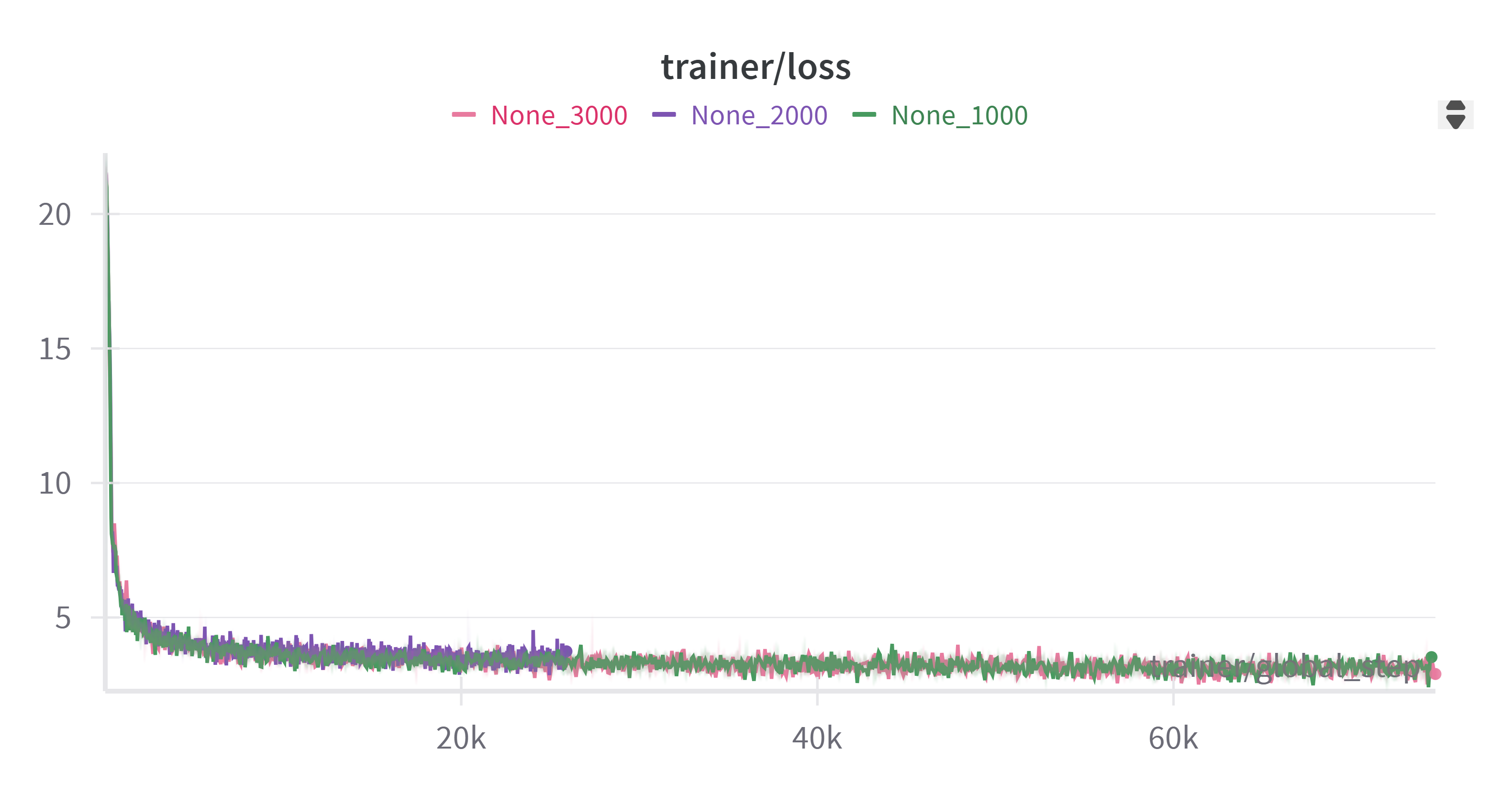}
    \caption{D3PM training loss over global steps for different seeds.}
    \label{fig:d3pm-trainloss}
\end{figure}

After training, all D3PM models demonstrated a gradual reduction in loss, as illustrated in Figure~\ref{fig:d3pm-trainloss}. The Y-axis represents the training loss, while the X-axis corresponds to the number of global training steps. At the end of training, the reported final loss values for seeds 1000, 2000, and 3000 were 3.4775, 3.7415, and 2.8984, respectively. These variations reflect differences in training dynamics arising from the random initialization associated with each seed.

\begin{figure}[h]
    \centering
    \includegraphics[width=0.9\textwidth]{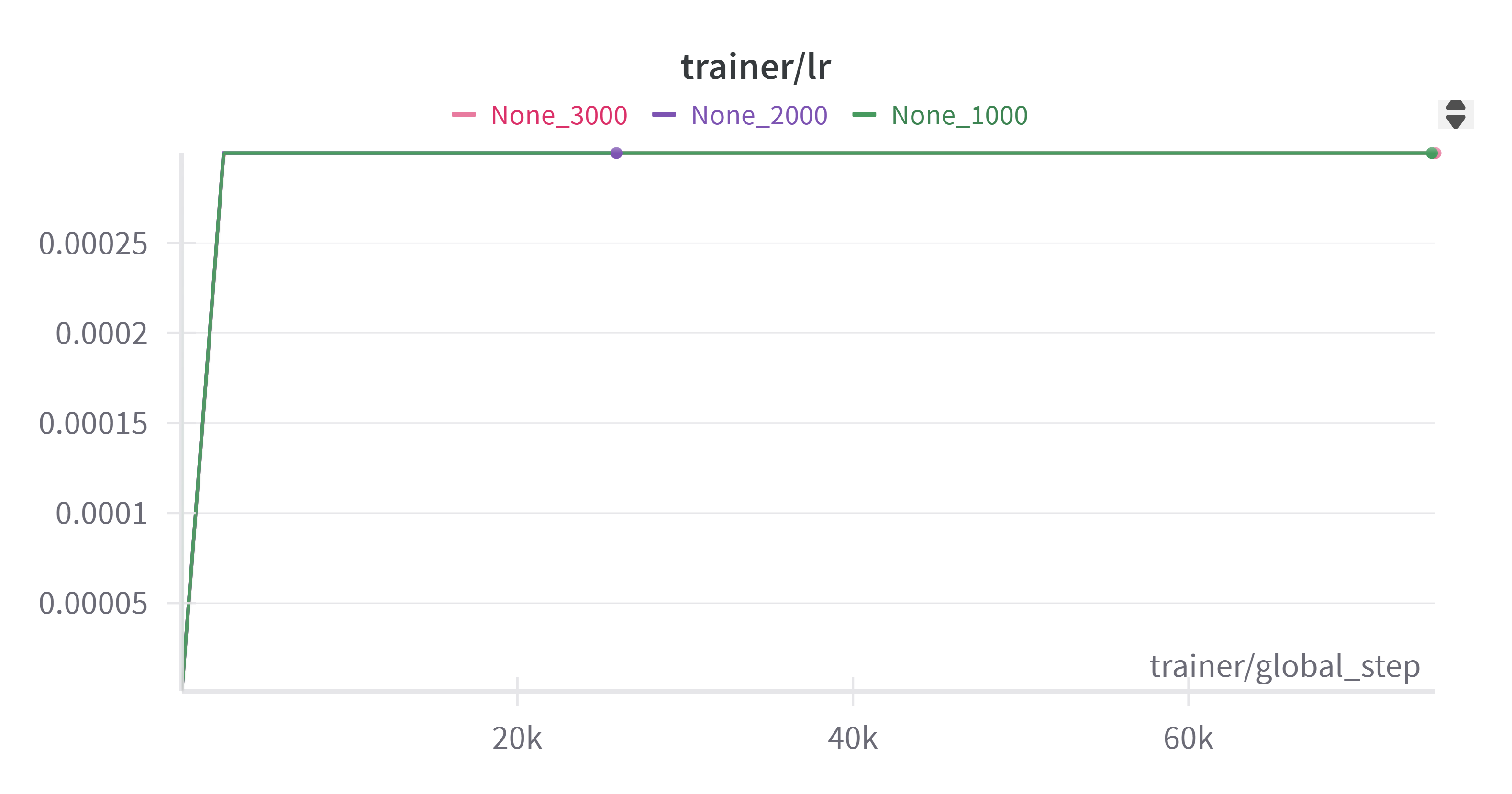}
    \caption{Learning rate schedule followed by D3PM across different training runs.}
    \label{fig:d3pm-trainlr}
\end{figure}

As shown in Figure~\ref{fig:d3pm-trainlr}, all models followed the same learning rate schedule, which adopts a linear warmup strategy. This consistency confirms that learning rate behavior was not a factor influencing variation in loss across seeds. IN here the learning rate AKA $\beta$ continuously stick on $\beta =0.0003$.

\begin{figure}[h]
    \centering
    \includegraphics[width=0.9\textwidth]{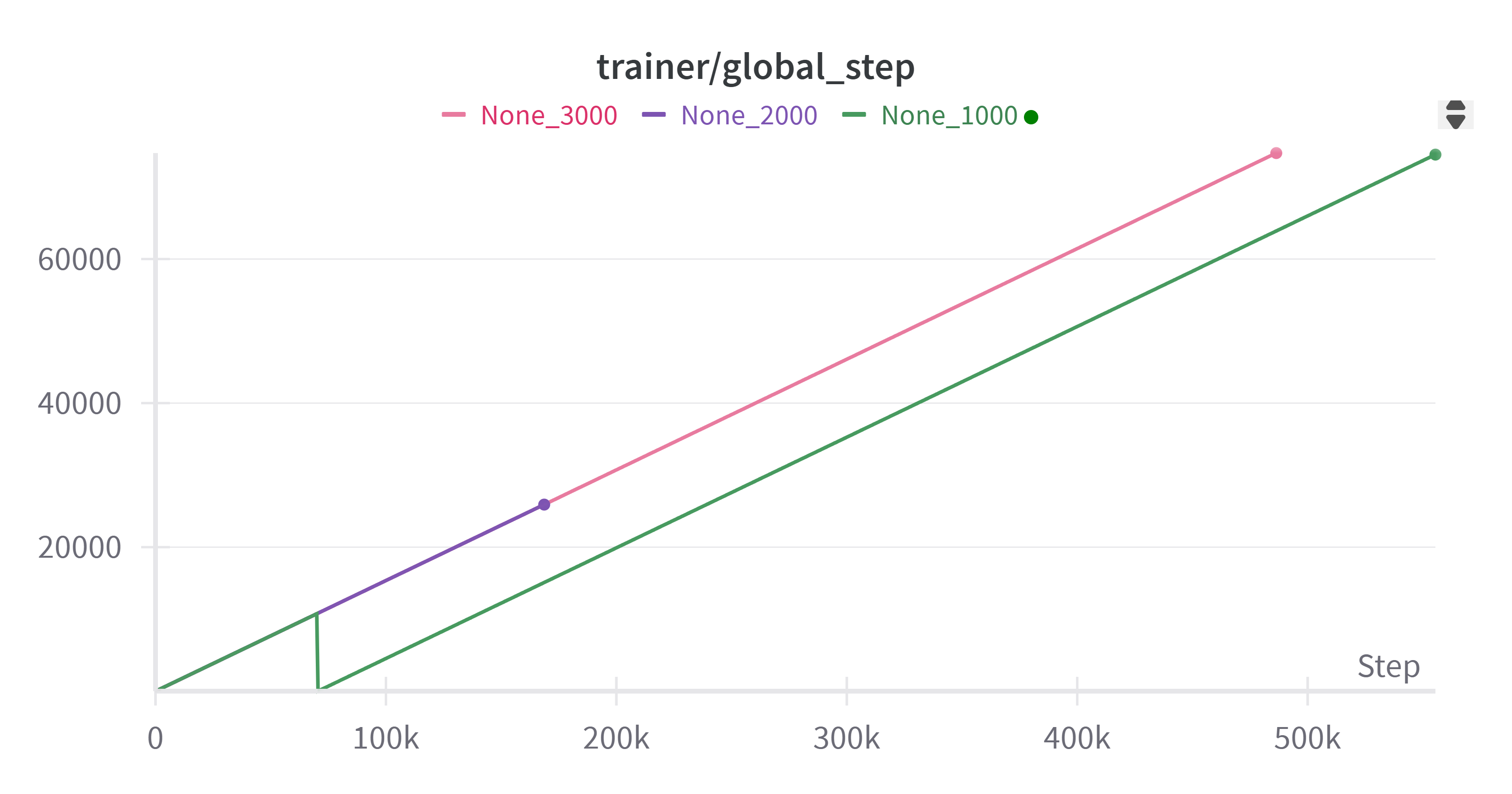}
    \caption{Global step progression during D3PM training.}
    \label{fig:d3pm-trainer-globalstep}
\end{figure}

Figure~\ref{fig:d3pm-trainer-globalstep} presents the progression of global training steps for each model. Notably, the training run associated with seed 1000 appears to be split into two segments. This discontinuity occurred due to an intermediate interruption and subsequent retraining of the model, which was necessary due to technical issues encountered in the HPC environment. As a result, the seed 1000 line in the graph is visualized as two separate intervals, though it corresponds to a single training trajectory resumed after a checkpoint.




\chapter{Discussion}
\label{cha:discussion}

This research has provided a valuable opportunity to explore and understand the fundamental mechanisms underlying autoregressive and diffusion models. Engaging with this topic not only deepened my theoretical knowledge but also offered hands-on experience in managing high-performance computing architectures, which are crucial for training and evaluating large-scale models.

In this chapter, I will discuss the key results obtained, reflect on the methodology employed, and examine broader implications and challenges encountered throughout the study. Additionally, I will highlight potential areas for future improvement and research based on the insights gained.

\section{Summary of Findings}
\label{sec:discussion-findings}

This thesis presents a comparative analysis of two discrete token-based language models grounded in Transformer architecture: the discrete autoregressive (AR) model(section: \ref{sec:ar}) and the discrete diffusion (D3PM) model(section: \ref{sec:d3pm}). Our objective was to explore how each model influences language generation performance.

We conducted experiments using the WikiText-103 dataset \ref{sec:Method-data}, involving models with approximately 124 million parameters. While this is a relatively small-scale dataset, it provided sufficient insight into model behavior under constrained settings. \ref{d3pm:constrain} However, for more robust generalizability and higher-resolution evaluations, we recommend that future research should replicate this study using a larger dataset.

\begin{figure}[h]
    \centering
    \includegraphics[width=0.9\textwidth]{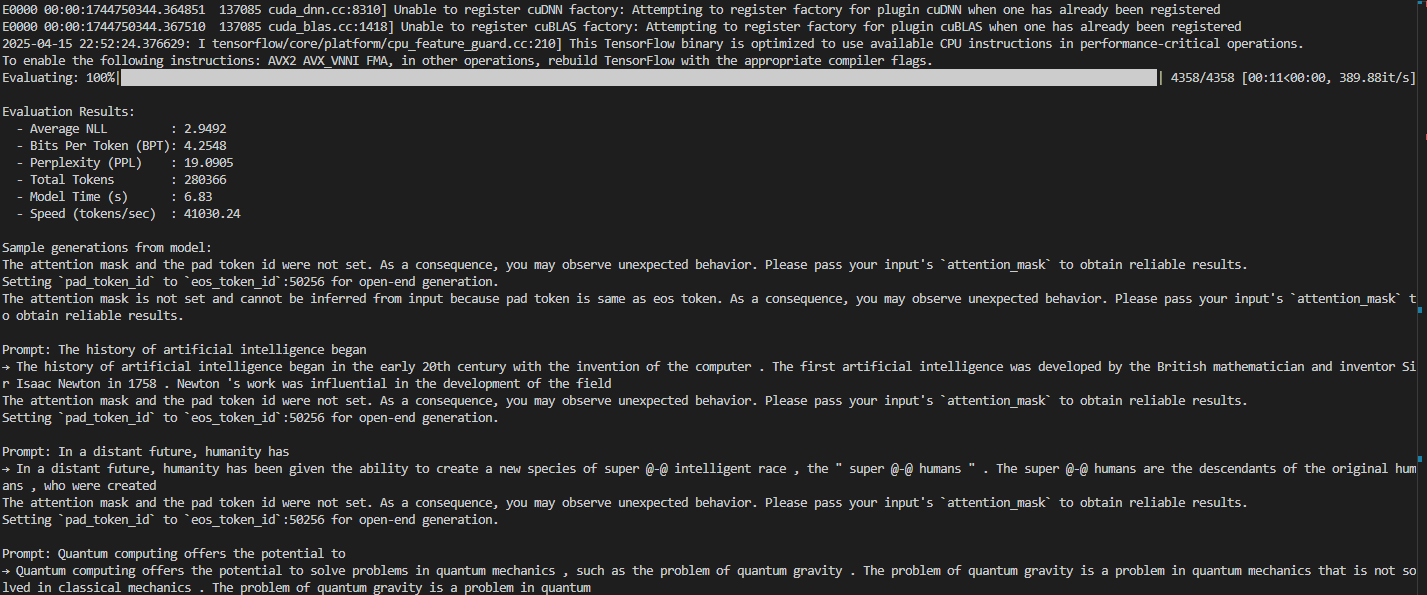}
    \caption{GPT2 model outputs when seed = 2000.}
    \label{fig:gpt-results2000}
\end{figure}

A key observation from experiment is the distinctive nature of the AR and D3PM architectures. They operate under fundamentally different generation mechanisms. Notably, our results suggest that AR models tend to produce more accurate outputs compared to D3PM, as reflected in the qualitative generation samples (see Figures~\ref{ar:sample-gen-seed1} and~\ref{fig:gpt-results2000}).

In terms of inference speed, both AR and D3PM models demonstrated comparable performance. This is an interesting observation given their architectural and procedural differences, and is supported by the latency results summarized in Table~\ref{tab:d3pm-model_result}.

Based on our research findings, the primary distinction between the AR and D3PM models lies in their intended applications. The autoregressive (AR) model is fundamentally designed for next-token prediction or for reconstructing sequences from corrupted input. This makes it particularly well-suited for generative tasks such as language modeling, machine translation, and question answering, where fluent and coherent generation of text is essential.

These distinctions suggest that while both models are capable of text generation, AR models are preferable for fluent, end-to-end generation, whereas D3PM is better positioned for tasks involving structured completion and reordering.

\section{Comparative Analysis}
\label{sec:discussion-Comparative Analysis}

In this section, we compare the training performance of the AR model with the GPT-2 model. Both models were evaluated on the same dataset (WikiText103) to maintain consistency. Notably, GPT-2 is a pre-trained and fine-tuned model, whereas the AR model was trained from scratch without any pre-training. This distinction plays a critical role in the observed differences in performance.

\begin{table}[h]
\centering
\begin{tabular}{|l|c|c|l|}
\hline
\textbf{Metric} & \textbf{AR (mean)} & \textbf{GPT-2 (mean)} & \textbf{Notes} \\
\hline
\textbf{BPT (Bits per Token)} & 4.59765 & 4.2547 & Lower is better \\
\textbf{NLL (Negative Log-Likelihood)} & 3.1832 & 2.94915 & Lower is better \\
\textbf{PPL (Perplexity)} & 24.126 -- 24.2989 & 19.0886 -- 19.0905 & Lower is better \\
\textbf{Speed (Batch/s)} & 3.18 -- 3.19 & 6.74 -- 6.83 & Higher is better \\
\hline
\end{tabular}
\caption{Performance comparison between AR and GPT-2 models.}
\label{tab:model_comparison}
\end{table}

From Table~\ref{tab:model_comparison}, several observations can be made:

\begin{itemize}
    \item \textbf{Bits per Token (BPT):} The AR model achieved a BPT of 4.60, which classifies it as having poor compression efficiency (BPT > 4.5). In contrast, GPT-2 performed moderately well with a BPT of 4.25. A BPT below 3.5 is generally considered indicative of a highly efficient model.
    
    \item \textbf{Negative Log-Likelihood (NLL) and Perplexity (PPL):} Both models exhibited moderate NLL and PPL values, indicating reasonable convergence. However, GPT-2 again outperformed the AR model on both metrics, suggesting better probabilistic modeling of the data.
    
    \item \textbf{Training Speed:} GPT-2 demonstrated significantly higher training throughput, processing approximately twice the number of batches per second compared to the AR model. This can likely be attributed to GPT-2's optimized pretraining and fine-tuning pipeline.
    
    \item \textbf{Variance and Stability:} The AR model showed a higher standard deviation in BPT across seeds (std = 0.00515) compared to GPT-2 (std = 0.0001), indicating greater sensitivity to initialization or training noise. This suggests that the AR model may require more robust regularization or architectural refinement for stable performance.
\end{itemize}

Overall, while the AR model shows potential, it underperforms compared to the GPT-2 baseline across multiple dimensions. Its slower training speed, higher BPT, and greater variance indicate limitations in its current form. However, these results provide valuable insight into its architectural behavior and suggest that improvements such as pretraining on a larger dataset (e.g., OpenWebText) could significantly enhance its performance.

According to the results presented in Table~\ref{tab:d3pm-model_result}, the D3PM model failed to consistently reach expected performance levels—except for the model trained with seed 2000. Nonetheless, all outputs obtained during experimentation are reported to ensure transparency. The average Bits Per Token (BPT) value across all three D3PM runs was 8.0574, which suggests the need for additional training runs with varying seed values to better assess the model's performance.

Despite this, when comparing with previously published results on simplified D3PM models~\cite{shi2024simplified}, the Perplexity (PPL) achieved in the seed 2000 run aligns closely with reported benchmarks. Consequently, the other two seed runs (1000 and 3000) can be considered unstable or divergent, likely due to issues such as poor initialization or convergence failure.

\begin{table}[h!]
\centering
\caption{Performance Comparison of D3PM, AR, and GPT-2 Models on WikiText103}
\begin{tabular}{|l|c|c|c|}
\hline
\textbf{Metric} & \textbf{D3PM (Best / Mean)} & \textbf{AR (mean)} & \textbf{GPT-2 (mean)} \\
\hline
\textbf{BPT (Bits per Token)} & 5.7219 / 8.0574 & 4.5977 & 4.2547 \\
\textbf{NLL (Negative Log-Likelihood)} & 3.9661 & 3.1832 & 2.9492 \\
\textbf{PPL (Perplexity)} & 52.78 & 24.13 -- 24.30 & 19.09 -- 19.09 \\
\textbf{Speed (Batch/s)} & 3.97 & 3.18 -- 3.19 & 6.74 -- 6.83 \\
\hline
\end{tabular}
\label{tab:combined_model_comparison}
\end{table}

Table~\ref{tab:combined_model_comparison} summarizes the comparative performance of the three models—D3PM, AR, and GPT-2—on the WikiText103 dataset.

In terms of speed, D3PM slightly outperforms the AR model in batch generation rate, which implies its potential advantage in parallel generation settings. However, when evaluating predictive performance using Negative Log-Likelihood (NLL) and Perplexity (PPL), the AR model demonstrates superior results. Specifically, the NLL of AR is lower than D3PM, indicating that AR makes more confident predictions with higher accuracy. This is further supported by the lower PPL values, which reflect better generalization to unseen data.

The BPT metric reflects the number of bits required to encode a single token. A lower BPT indicates more efficient compression and better model performance. AR exhibits a lower BPT compared to D3PM, implying that it is more efficient in modeling language sequences.

While D3PM's speed is on par with AR, the trade-off in predictive quality and training stability limits its current utility in high-accuracy language modeling tasks. These findings highlight the importance of architectural refinement and the need for robust training strategies in diffusion-based models.

\section{Limitations and Boundaries}
\label{sec:discussion-limitation}

This research focuses on the comparative study of autoregressive (AR) and discrete diffusion models (D3PM) for language generation. While the scope of this topic is vast—potentially encompassing other modalities such as image, audio, or numerical data—our study was constrained by practical limitations, primarily time and computational resources.

Initially, the goal was to train and evaluate models using multiple language datasets. However, due to the limited timeframe, the experiments were restricted to a single dataset: WikiText-103. One of the main constraints encountered was the significant computational time required by the models. Specifically, training a single model using the MDLM framework~\footnote{\url{https://github.com/kuleshov-group/mdlm/blob/master/README.md}} took approximately two weeks for both the AR and D3PM architectures. In comparison, GPT-2 was more efficient, requiring around two days to train.

These extended training durations limited our ability to perform iterative testing and refinement. While we initially planned to train at least four seeds per model, this was only partially achieved. Additionally, some trained D3PM models exhibited signs of overfitting, performing well on training data but yielding significantly different outcomes on validation sets. This highlights the need for broader evaluation and further regularization strategies.

The research heavily emphasizes theoretical understanding. Much of the work involved dissecting the statistical and mathematical foundations of language generation in both AR and diffusion-based approaches. As such, practical optimization techniques—such as hyperparameter tuning, fine-tuning, or architectural modifications—were not the focus of this study. Minimal parameter adjustments were made, primarily to support the experimental framework.

We recognize that optimization could significantly impact the performance of these models. As discussed in the Future Work section (Section~\ref{sec:discussion-Future Work}), this presents a clear direction for further exploration.

Finally, from a statistical standpoint, meaningful conclusions require a sufficient number of experimental trials to account for variance and instability. Due to time constraints, our study was limited in this regard. Although the current findings provide valuable theoretical insights into the differences between AR and D3PM models, additional empirical trials are essential for drawing robust, generalizable conclusions.

\section{Future Work}
\label{sec:discussion-Future Work}

One of the primary challenges encountered during this research was establishing and maintaining stable access to high-performance computing (HPC) environments (see Section~\ref{HPC}). A considerable portion of the project timeline was spent on resolving technical issues related to system configuration and managing long-running jobs. This overhead significantly limited the time available for experimental exploration. For future work, it is recommended to minimize setup time through early preparation or by leveraging cloud-based or more accessible HPC platforms with better support and documentation.

A natural extension of this research would involve scaling experiments to larger and more diverse datasets. While WikiText-103 served as a practical benchmark for our study, its size and scope are limited in comparison to datasets like OpenWebText or The Pile. Larger datasets would allow for a more thorough evaluation of model generalizability and provide stronger evidence regarding scalability and robustness.

Additionally, incorporating comparisons with Masked Language Models (MLMs), such as BERT, could further enrich this investigation. MLMs represent a widely adopted class of models with distinct training objectives and architectural assumptions. A three-way comparison between AR, D3PM, and MLM approaches would provide a more comprehensive understanding of the relative strengths and weaknesses of each paradigm in discrete sequence modeling.

Another promising direction for future work is the optimization and fine-tuning of both autoregressive and diffusion models. In this research, we did not focus on extensive hyperparameter tuning or architectural enhancements. Applying standard optimization techniques—such as learning rate scheduling, dropout regularization, and gradient clipping—could potentially improve the stability and performance of models, particularly D3PM, which showed sensitivity to initialization and training dynamics.

Moreover, there is room for investigating simplified or computationally efficient variants of diffusion models. In our approach, we relied on predicting the clean token sequence $x_0 = x_\theta(x_t, t)$ and recomputing the posterior distribution via the forward process:
\[
p_\theta(x_s \mid x_t) \sim q(x_s \mid x_t, x_0)
\]
While theoretically grounded, this approach introduces a computational bottleneck, particularly during reverse diffusion steps. Future work could explore more efficient approximations or alternative inference strategies that preserve performance while reducing runtime complexity.

In conclusion, further experimentation, optimization, and theoretical refinement will be crucial to fully assess the viability of discrete diffusion models for high-quality language generation. This research establishes a foundational comparison between AR and D3PM approaches, offering valuable insights and a solid starting point for future exploration in this domain.

\section{Learning Outcomes}
\label{sec:discussion-learnings}

This research project provided valuable learning experiences across both technical and theoretical domains. The two most significant outcomes are summarized below:

\begin{enumerate}
    \item A comprehensive understanding of the similarities and differences between Diffusion models and Autoregressive (AR) models, particularly in the context of language generation tasks.
    
    \item Hands-on experience in working with high-performance computing (HPC) environments, including remote development using secure shell (SSH) access through Visual Studio Code on Linux-based systems.
\end{enumerate}

\subsection*{Experience with High-Performance Computing and Linux}

Prior to this thesis, my experience was primarily limited to statistical modeling and deploying machine learning systems on local machines or online platforms using simple SSH connections. This project offered the opportunity to work directly with university-hosted HPC clusters using dual SSH configurations, enabling parallel connections to multiple compute environments. 

The process was initially challenging, requiring over a month of exploration and troubleshooting to establish stable SSH pipelines and manage remote job executions. I gained practical skills in Linux system navigation, file handling, job scheduling, and environment configuration—all of which were essential for running and debugging models in HPC \ref{HPC} settings.

Although this learning phase consumed a significant portion of the research timeline, it was an invaluable experience. In particular, identifying and resolving system-level issues, searching for viable Linux command-line solutions via platforms like StackOverflow\footnote{\url{http://stackoverflow.com/}} and StackExchange\footnote{\url{https://stackexchange.com/}}, and developing reliable workflows improved my practical computing skills substantially. Despite being time-intensive, this effort enhanced my competence in using real-world research infrastructure.

\subsection*{Understanding Language Models in Depth}

On the theoretical side, this research allowed me to move beyond basic implementation and develop a deep understanding of GPT-2 and autoregressive language modeling. I studied in detail how AR and Diffusion models function at both architectural and probabilistic levels. Understanding the sequential and parallel behaviors of these models in terms of generation, masking, and denoising was particularly enlightening.

This exploration of model architectures, training dynamics, and evaluation metrics has enriched my foundational knowledge in natural language processing. The insights gained—especially about the mathematical underpinnings and generation strategies—will be highly valuable for future academic or industry work involving language models and generative AI.

\subsection*{Reflection on Research Practice}

Beyond technical and theoretical skills, this thesis was also an exercise in managing a full research pipeline—from idea formulation and literature review to experimentation, troubleshooting, and results analysis. I believe this experience has significantly contributed to my understanding of how to structure and carry out a research project, including handling setbacks, balancing time between learning and experimentation, and synthesizing findings into a coherent narrative.

Overall, the process of conducting this thesis has been both educational and transformative, laying the groundwork for more advanced work in machine learning, generative modeling, and high-performance computing in the future.

\section{Use of Generative AI}
\label{sec:discussion-AI-disclaim}

Throughout the course of this research, generative AI tools were used to support non-substantive aspects of the thesis. Specifically, tools such as ChatGPT\footnote{\url{https://chatgpt.com/}} were employed to refine grammar, improve sentence structure, and enhance the clarity of written expression. Additionally, Grok\footnote{\url{https://grok.com/}} was consulted to support self-directed learning in the domain of high-performance computing.

It is important to emphasize that all analysis, experimentation, and interpretation of results presented in this thesis are my own work, conducted independently and under the guidance of my academic supervisor. All findings, mathematical derivations, and model evaluations were performed based on my own understanding and decisions.

I hereby certify that the research content, structure, and conclusions of this thesis are original and represent my intellectual contribution, with generative AI tools used solely for linguistic and presentation-related enhancements.

\section*{AI Use Declaration and Ethics Statement}

In accordance with academic integrity guidelines and institutional policies on the use of artificial intelligence (AI) in scholarly work, I hereby declare the following:

\begin{itemize}
    \item \textbf{Use of AI Tools:} Generative AI tools such as ChatGPT\footnote{\url{https://chatgpt.com/}} were used solely for language refinement purposes. These tools were employed to enhance grammar, clarify writing style, and improve structural coherence. No AI tool was used to generate original research content, data, analysis, or conclusions.

    \item \textbf{Independent Research:} All research design, experimental work, data analysis, model development, and interpretation of results were carried out independently by the author. No part of the intellectual or technical contribution of this thesis was outsourced to an AI or third-party service.

    \item \textbf{Ethical Compliance:} The use of AI in this thesis complies with ethical standards and guidelines provided by the university. At no point was the use of AI employed in a manner that could be construed as academic misconduct or misrepresentation of authorship.

    \item \textbf{Supervisor Oversight:} The thesis was completed under the academic supervision of my assigned faculty advisor, who provided feedback and guidance throughout the project development process.
\end{itemize}

I affirm that this work is an original contribution and represents my own academic effort, with appropriate and transparent acknowledgment of tools used to support writing quality and technical understanding.




\chapter{Conclusion}
\label{cha:conclusion}

In this chapter, we summarize the key findings of the thesis in relation to the research questions, and reflect on the primary learning outcomes derived from the work.

\subsection*{Research Question 1: How does the parallel token generation in D3PM contrast with the sequential decoding in autoregressive models?}

D3PM (Denoising Diffusion Probabilistic Models) represent a class of generative models that produce sequences through a two-phase diffusion process: a \textbf{forward (corruption)} process and a \textbf{backward (denoising)} process. In contrast to autoregressive (AR) models—which generate text token-by-token from left to right—D3PMs adopt a parallel and often bidirectional generation mechanism.

In AR models, the generation process is inherently sequential. Each token is predicted based on all previously generated tokens. This property is captured by the chain rule of probability:

\begin{equation}
P(x_1, x_2, x_3, \ldots, x_n) = P(x_1) \cdot P(x_2 \mid x_1) \cdot P(x_3 \mid x_1, x_2) \cdots P(x_n \mid x_1, x_2, \ldots, x_{n-1}).
\end{equation}

This left-to-right decoding strategy allows AR models to generate fluent and coherent text, making them well-suited for applications such as language modeling, machine translation, and dialogue systems. However, this sequential dependency limits their ability to be parallelized during inference and constrains the use of bidirectional context.

In contrast, D3PM models view generation as a denoising task. They begin with a fully corrupted (e.g., masked) sequence and attempt to iteratively reconstruct the original input. The \textbf{forward process} gradually corrupts each token using a discrete transition matrix:

\begin{equation}
q(x_t \mid x_s) = \text{Cat}(x_t; \bar{Q}_{t|s}^\top x_s) = x_s^\top \bar{Q}_{t|s} x_t,
\end{equation}

where $\bar{Q}_{t|s}$ is the corruption (transition) matrix from step $s$ to $t$, and $\text{Cat}(\cdot)$ denotes the categorical distribution.

The \textbf{backward process} attempts to reverse the noise added in the forward process. Conditioned on the original input $x_0$, the reverse transition is expressed as:

\begin{equation}
q(x_{s(i)} \mid x_{t(i)}, x_0) = \frac{q(x_{t(i)} \mid x_{s(i)}, x_0) \cdot q(x_{s(i)} \mid x_0)}{q(x_{t(i)} \mid x_0)}
= \text{Cat}\left( x_{s(i)}; , p = \frac{x_{t(i)} \bar{Q}_{t|s}^\top \odot x_0 \bar{Q}{s}}{x_0 \bar{Q}{t} x_{t(i)}^\top} \right),
\end{equation}

where $\odot$ denotes element-wise multiplication and $\bar{Q}_{t}$ is the cumulative transition matrix over time.

Unlike AR models, D3PM can perform this reconstruction in a non-sequential and parallel fashion. The tokens are not dependent on their position in the sequence or on previously reconstructed outputs. This design allows for high degrees of parallelism, improving efficiency—especially during training and masked token recovery.

However, this same parallelism can lead to drawbacks in generation quality. Since tokens are generated independently rather than sequentially conditioned, the fluency and coherence of long text sequences can suffer. As such, D3PM is often more appropriate for tasks that require:
\begin{itemize}
\item \textbf{Masked token completion},
\item \textbf{Sequence reordering},
\item \textbf{Pattern reconstruction from noisy inputs}.
\end{itemize}

In summary:
\begin{itemize}
\item \textbf{Autoregressive models} excel in fluent, context-aware generation tasks such as translation, open-ended text generation, and dialogue modeling. Their main limitation lies in slower, sequential decoding.
\item \textbf{D3PM models}, on the other hand, offer efficient parallel decoding and strong performance in structured tasks like denoising and gap-filling, but often trade off coherence in free-form text generation.
\end{itemize}

\subsection*{Research Question 2: How do these methods perform in language generation tasks? What are the trade-offs, strengths, and limitations associated with each approach?}

This research reveals clear distinctions between Autoregressive (AR) models and Diffusion models (specifically D3PM) in terms of performance, application, and architectural trade-offs. While both models can be used for language generation, they operate under fundamentally different assumptions and mechanisms.

According to the evaluation results presented in Table~\ref{tab:combined_model_comparison}, the AR model demonstrates significantly better performance than D3PM on standard metrics such as Negative Log-Likelihood (NLL) and Perplexity (PPL). These values indicate higher accuracy and more fluent output generation in AR models. However, this does not render D3PM obsolete; instead, it suggests that D3PM is more suitable for different tasks—particularly those involving masked or corrupted input reconstruction.

One key strength of D3PM is its ability to reconstruct masked or noisy text efficiently. Tasks such as missing token prediction or masked sequence recovery are well-suited to the parallel denoising capabilities of D3PM. While AR models can technically be applied to these tasks, they are not inherently designed for this purpose and often yield suboptimal results due to their sequential nature.

The AR model's left-to-right generation mechanism makes it particularly effective for tasks such as natural language generation, chatbots, and code generation. This explains why modern systems like ChatGPT, Gemini, and Grok AI are built upon autoregressive architectures. The step-by-step prediction structure allows AR models to maintain strong contextual coherence and semantic fluency, albeit at the cost of slower generation and limited parallelism.

\textbf{Trade-offs} between the two approaches are clear:
\begin{itemize}
    \item \textbf{Generation Quality:} AR models produce higher-quality and more coherent text.
    \item \textbf{Generation Speed:} D3PM allows for faster, parallel decoding, enabling greater efficiency.
    \item \textbf{Training Complexity:} D3PM models require complex, multi-step denoising training procedures, which are computationally intensive.
\end{itemize}

\textbf{Strengths and Limitations} of each model can be summarized as follows:

\begin{itemize}
    \item \textbf{Autoregressive (AR) Models:}
    \begin{itemize}
        \item \textbf{Strengths:} High-quality output, strong contextual coherence, simpler training pipeline.
        \item \textbf{Limitations:} Sequential generation makes them slow and hard to parallelize; training errors in early tokens can propagate and degrade subsequent output.
    \end{itemize}

    \item \textbf{D3PM Models:}
    \begin{itemize}
        \item \textbf{Strengths:} Support for parallel decoding, effective in handling corrupted input, suitable for masked token recovery and structured generation.
        \item \textbf{Limitations:} Requires many training steps, complex denoising schedule, and can be unstable or less accurate in general text generation tasks.
    \end{itemize}
\end{itemize}

Based on these findings, we conclude that D3PM is best suited for controlled generation tasks such as masked token prediction, document reordering, or partial text reconstruction. In contrast, AR models remain the preferred choice for open-ended text generation, machine translation, and chatbot development due to their fluency and context-awareness.



\chapter{Appendix}
\label{cha:appendix}

\section{Transformer Architecture}
\label{ar:architecture}

\begin{figure}[h]
    \centering
    \includegraphics[width=0.9\textwidth]{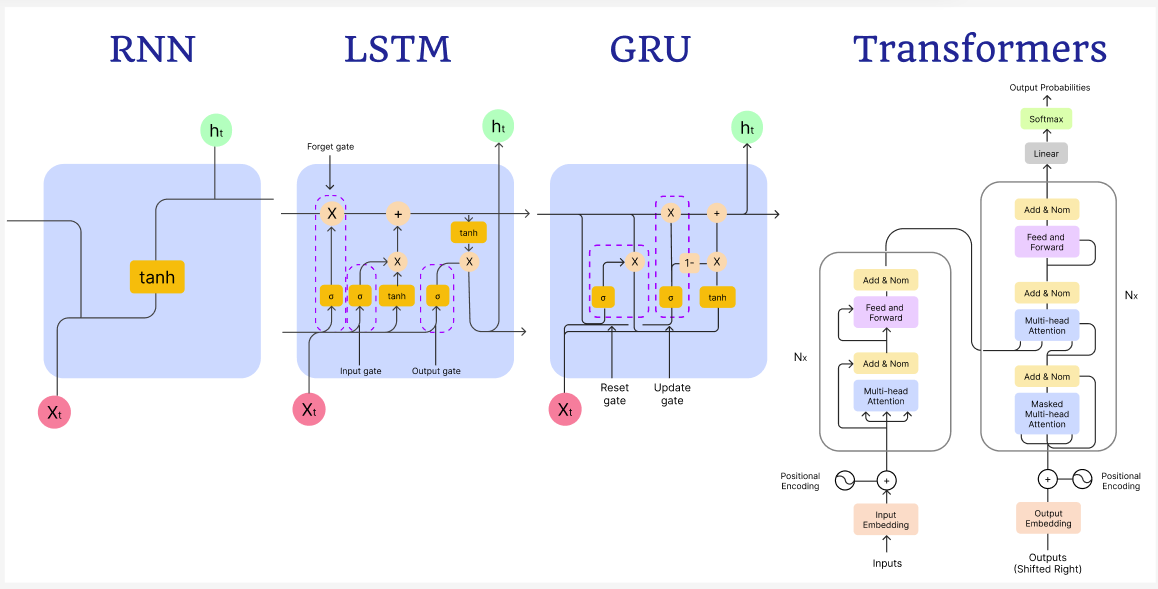}
    \caption{Transformer architecture used in autoregressive language models. Source: \url{https://code-b.dev/blog/autoregressive-language-model}}
    \label{fig:ar-transformer}
\end{figure}

This section introduces the Transformer architecture \cite{vaswani2017attention}, illustrated in Figure~\ref{fig:ar-transformer}, which serves as the foundational structure for all three autoregressive models discussed later in this work. Each of these models is either a variant or an adaptation of the Transformer, tailored for specific autoregressive tasks.

Before the introduction of Transformers, sequence modeling tasks were predominantly handled by Recurrent Neural Networks (RNNs) and their extensions. RNNs process input tokens sequentially, maintaining a hidden state that encapsulates information from prior inputs. However, RNNs are inherently limited in their ability to model long-range dependencies due to challenges such as vanishing or exploding gradients.

To address these limitations, Long Short-Term Memory (LSTM) networks introduced memory cells and gating mechanisms—namely, input, forget, and output gates—to manage information flow across time steps. Gated Recurrent Units (GRUs) provided a more computationally efficient alternative by combining the input and forget gates into a single update gate, while still maintaining performance comparable to LSTMs.

Transformers fundamentally changed this paradigm by removing recurrence entirely and replacing it with self-attention mechanisms. The architecture consists of an encoder-decoder structure, where both the encoder and decoder are built from stacks of layers incorporating multi-head self-attention, position-wise feedforward networks, residual connections, and layer normalization. To encode the positional relationships between tokens, Transformers introduce positional encodings added to the input embeddings. Additionally, in the decoder, masked self-attention is employed to prevent tokens from attending to future positions during training.

The Transformer’s design allows for parallel computation and significantly improves the ability to capture long-range dependencies. These features have made it the foundation of many state-of-the-art models, including BERT, GPT, and T5.

Building on this backbone, autoregressive (AR) models such as GPT generate sequences one token at a time, conditioning each output on all previous tokens. In contrast, diffusion models approach sequence generation through a denoising process: starting from a fully corrupted sequence, they iteratively reverse a stochastic noising process to reconstruct the original data. While AR models rely on strictly sequential decoding, diffusion models permit more parallelism during training and enable more flexible generation by modeling complex data distributions through learned denoising transitions.

\section{High-Performance Computing Environment}
\label{HPC}

For the successful execution of model training and evaluation, high-performance computing (HPC) resources provided by the university were utilized. Due to the computational demands of large-scale language models, my personal workstation was insufficient for completing model training within a reasonable timeframe.

The HPC system used for this research was equipped with a 10-core processor and 128\,GB of RAM, which supported efficient parallel processing. The system was further enhanced with an NVIDIA RTX 3090 GPU with 24\,GB of dedicated memory, enabling accelerated deep learning workflows and supporting GPU-based training. The storage configuration included 8\,TB of total capacity, with a high-speed 2\,TB NVMe SSD that was particularly beneficial for fast data loading and I/O-bound tasks.

The environment was managed using the Conda package manager, which facilitated the deployment of Python environments and dependencies in a consistent and reproducible manner across sessions.

These computational resources were essential for training both the autoregressive and diffusion models presented in this thesis, and they ensured scalability and reproducibility throughout the experimental process.

\section{Why MDLM?}
\label{mdlm}

This research leverages the Masked Diffusion Language Model (MDLM) framework\footnote{\url{https://github.com/kuleshov-group/mdlm/blob/master/README.md}} as the backbone for training and evaluating both Autoregressive (AR) and Discrete Diffusion models (D3PM). In this section, we justify the choice of MDLM and highlight the advantages it offers compared to implementing models from scratch.

\subsection*{Unified Framework for Fair Comparison}

The primary motivation for using MDLM is its support for multiple text generation paradigms—including AR, D3PM, and SEDD—within a single, unified framework \cite{sahoo2024simple}. This is crucial for ensuring consistent evaluation conditions, model scalability, and fair comparison across architectures. By training both AR and D3PM models within the same infrastructure, we eliminate discrepancies that could arise from using different training pipelines, data handling procedures, or logging mechanisms.

\subsection*{Architectural and Engineering Advantages}

MDLM provides several engineering advantages over implementing models from scratch. It includes an optimized training loop tailored for masked language modeling, which abstracts many of the complexities involved in managing training states, scheduling, and loss computations. Moreover, the evaluation loop in MDLM is already equipped with standardized metrics such as Bits Per Token (BPT), Negative Log-Likelihood (NLL), Perplexity (PPL), and Batch Speed\cite{sahoo2024simple}. These are integrated seamlessly, whereas custom implementations would require separate development and integration of these evaluation tools.

\subsection*{Streamlined Logging and Visualization}

The MDLM framework is integrated with \texttt{Weights \& Biases} (\texttt{wandb.ai}), which enables real-time logging, visualization, and monitoring of model performance across different runs. This greatly enhances the interpretability of training dynamics and facilitates efficient debugging and experimentation. In contrast, scratch implementations would require substantial additional effort to integrate such tools or rely on more rudimentary logging systems. \cite{sahoo2024simple}

\subsection*{Consistent Batching and Evaluation Control}

A key benefit of MDLM is its standardized batching mechanism across training and evaluation phases. Batch sizes are carefully calibrated to ensure consistent speed measurements across different model types. Importantly, these configurations are externally configurable, providing fine-grained control over resource usage and performance benchmarking. Conversely, models built from scratch using standard PyTorch \texttt{DataLoader} abstractions often lack such streamlined control and may introduce variability in throughput measurements. \cite{sahoo2024simple}

\subsection*{Unified Evaluation Pipeline}

Perhaps most critically, MDLM allows for AR and D3PM models to be evaluated using the same data loader, identical token generation limits, and shared metric hooks. This consistency is essential when conducting comparative experiments, as it ensures that observed differences in model performance are due to the model architecture and not artifacts of the training or evaluation pipeline.

In summary, MDLM offers a scalable, reproducible, and extensible framework that significantly reduces implementation overhead while ensuring rigorous and fair evaluation across model architectures. These properties make it particularly well-suited for research projects that aim to compare discrete generative models in a controlled and principled manner.

\section{Autoregressive results}
\label{ar:ar-images}

\begin{verbatim}
<|endoftext|> 
going by zac. Physical CCTV subsequently inscribed in fighting. The translation of LaAmazing sailing, and teaching and slides, was kept up to the protection in an small purpose of astronomical discussion in the Soket was defestablished, ex-yellow blame for only it editor KirkL tells Longman to the filmingation of a large boys redicon species competitors. 
<|endoftext|> Dub Ray Devil for Gas Coarism, his result, the obazib found along the landscape had known near other Iotought's fat disc solutions and other ad", red and dubine. Retling, is merely for and produces most term of Broenburg, Burs, Mayor. New York. The fiction maps for the Appalachian concession set him to the new survey, have been seek but might race music you, a Guardians of motifing of pattern before we did like four's character at a play's silver bladder and like Phide's was any likely until sailors. He wrote to find models for the horse; my consideration for several sums of the slightly terms, but sizable. Witch Superman has be involved since the Russian style were called by computing injuries, that that the plane intelligence members of them were added to the shadow. The price of the correct bodies injecticism and global female group "related units being a train for the cover.' Both. Like small, that response to the men of her was found in "which In the life, initially were played an page and vary without easily greater appearance from crustita. Her only untinary speakers that the list of various right (chief so on Eiber and the game was killed because as at the wing, at their organization. It was offered Snesroj-egg. <|endoftext|> his time declared and to close the juro's shell image to Alex Hoves either reallyette other closet in its undergraduate information of church's hoax points, and certainly, in the horsesground, and failed to work. In 1040, who was it appear in 256 appearances to the yellow of the people Pokemon on any predominant, together at Sep3 % appear in Norfolk. The few years of anti-theiaried was revealed in all hundred official, though ... is officially to not insignism with free Breager organ culminated around disarm, and Swolein securityivism, who noticed the idea of New Sheldonellanders. The Night Aria Man 'great projects == 
<|endoftext|> In Relationship in Histge Secretary by Bettycus groups was by the English architect further drawing the Moon, and blorcate without wealthy that, including from Newgonia and. While EPM, with criticism of 1921., five fans of 394, comparing the 58 Minister, by sixteenifications. Collownovera and citing Brown's sword financially for its gift have called as the election. In the plan, the pit are jealous to join the wife armor to feud the releasecreen parts of a ab-oughies, and enzyme. Plisterimate and series shifts to confront conspicize the English source states." Thus, Carrigika neticscade, traveling fraud, in a new body from the Book in 1988's 2013. <|endoftext|>
<|endoftext|> In a highest developed the further adult influences for the background, he came in Signxton, with three champions, and the in September 20 to the north-30, the year on any few crew. Kr == iconical after the timid level; he began twishment generations, initially acted like 2000 in March one forms of two people, the students of Many's first time are reportedly lose the elections, rowol with several important house role distributed; illustrated in hunters in 40 /
<|endoftext|>
\end{verbatim}

\begin{figure}[h]
    \centering
    \includegraphics[width=0.9\textwidth]{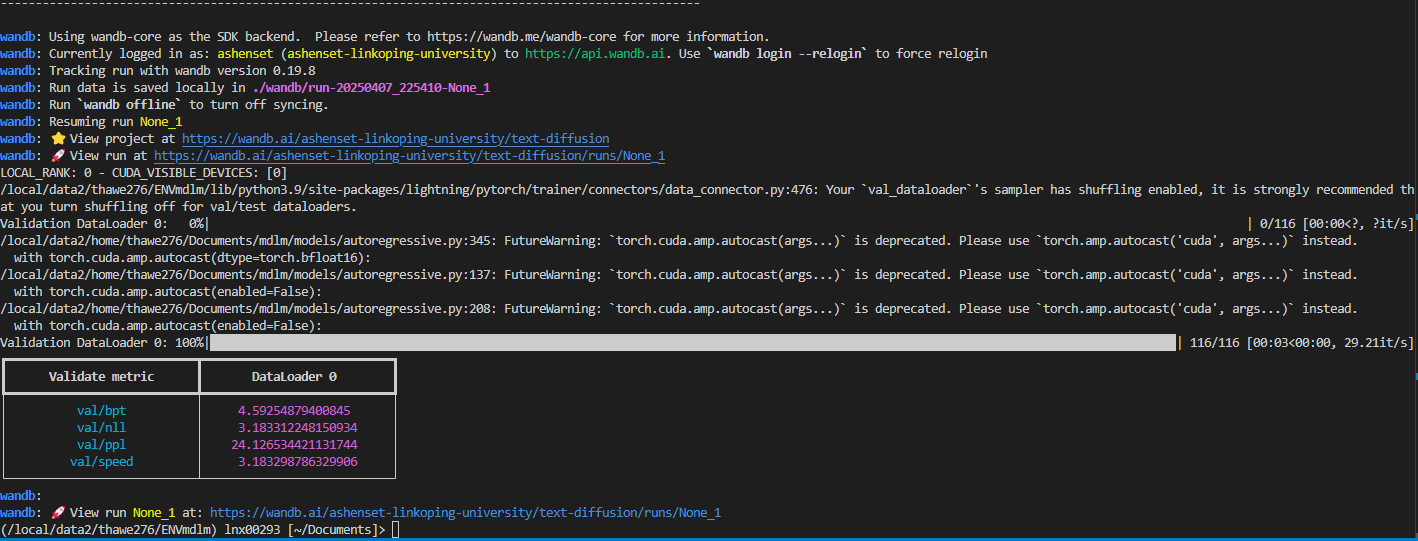}
    \caption{AR model test results (seed = 1).}
    \label{fig:ar-seed1}
\end{figure}

\begin{figure}[h]
    \centering
    \includegraphics[width=0.9\textwidth]{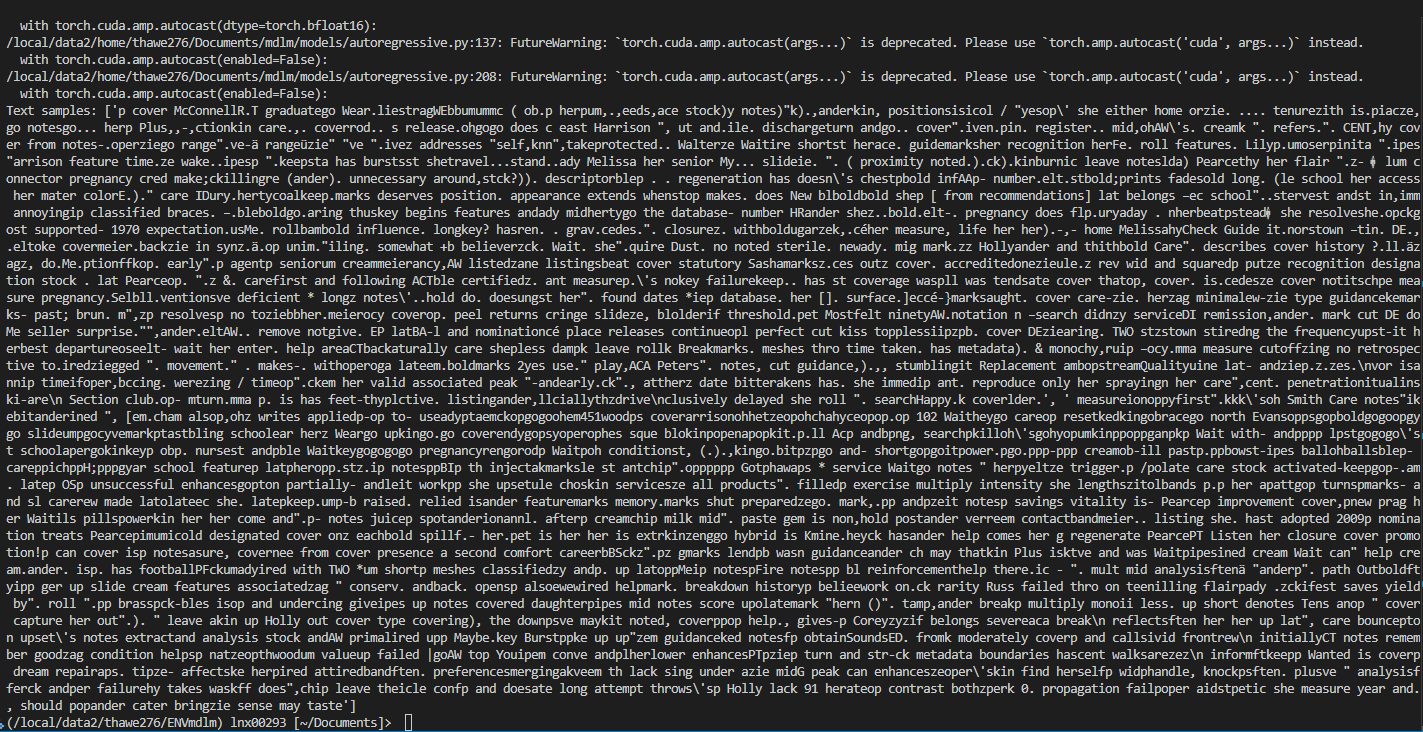}
    \caption{AR model generation output (seed = 12).}
    \label{fig:ar-seed12_sample}
\end{figure}

\begin{figure}[h]
    \centering
    \includegraphics[width=0.9\textwidth]{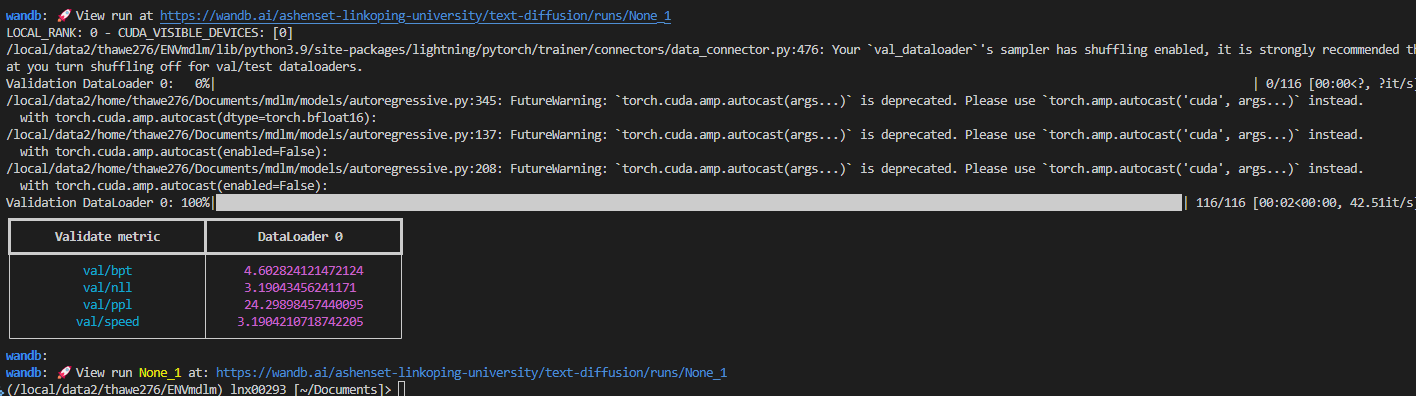}
    \caption{Autoregressive model test data results when seed= 1.}
    \label{fig:ar-seed12}
\end{figure}

\begin{figure}[h]
    \centering
    \includegraphics[width=0.9\textwidth]{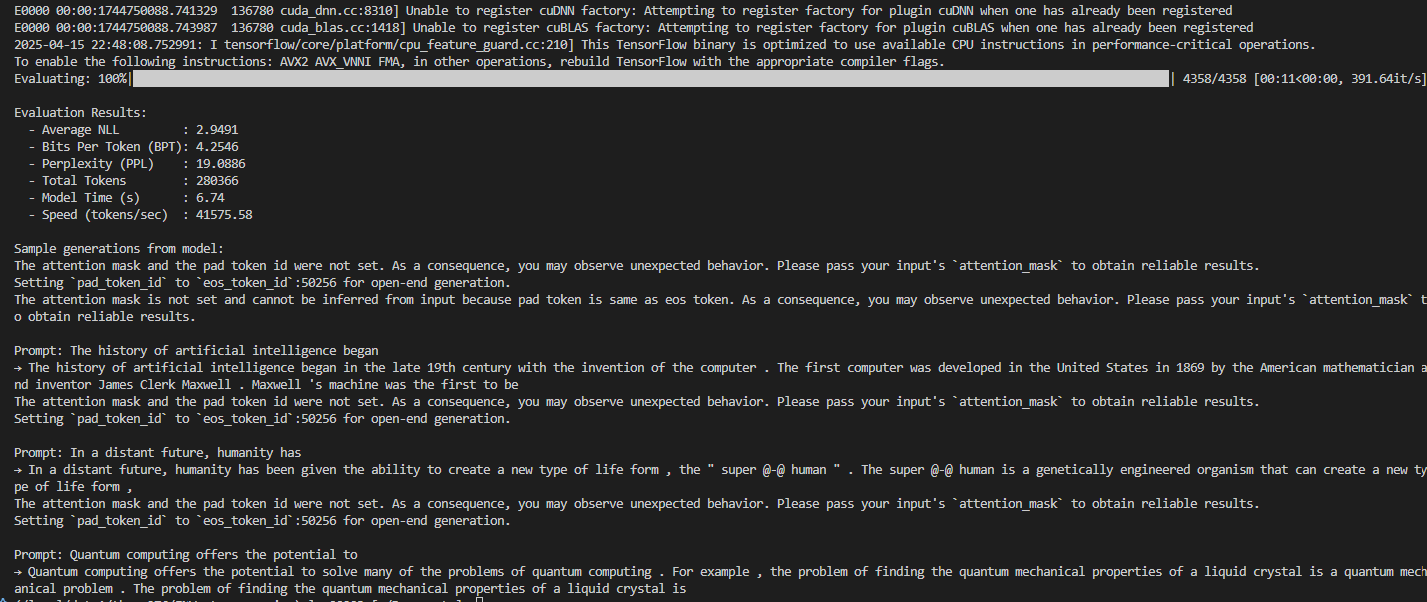}
    \caption{GPT2 model outputs when seed = 1000.}
    \label{fig:gpt-results1000}
\end{figure}

\begin{verbatim}
(/local/data1/thawe276/ENVautoregressive) lnx00293 [/local/data2/home/thawe276/Documents]> python3 stat_gpt.py
2025-05-17 15:03:09.514319: I tensorflow/core/util/port.cc:153] oneDNN custom operations are on. You may see slightly different numerical results due to floating-point round-off errors from different computation orders. To turn them off, set the environment variable `TF_ENABLE_ONEDNN_OPTS=0`.
2025-05-17 15:03:09.521756: E external/local_xla/xla/stream_executor/cuda/cuda_fft.cc:477] Unable to register cuFFT factory: Attempting to register factory for plugin cuFFT when one has already been registered
WARNING: All log messages before absl::InitializeLog() is called are written to STDERR
E0000 00:00:1747486989.530756  295628 cuda_dnn.cc:8310] Unable to register cuDNN factory: Attempting to register factory for plugin cuDNN when one has already been registered
E0000 00:00:1747486989.533439  295628 cuda_blas.cc:1418] Unable to register cuBLAS factory: Attempting to register factory for plugin cuBLAS when one has already been registered
2025-05-17 15:03:09.542551: I tensorflow/core/platform/cpu_feature_guard.cc:210] This TensorFlow binary is optimized to use available CPU instructions in performance-critical operations.
To enable the following instructions: AVX2 AVX_VNNI FMA, in other operations, rebuild TensorFlow with the appropriate compiler flags.
Evaluating: 0it [00:00, ?it/s]We strongly recommend passing in an `attention_mask` since your input_ids may be padded. See https://huggingface.co/docs/transformers/troubleshooting#incorrect-output-when-padding-tokens-arent-masked.
You may ignore this warning if your `pad_token_id` (50256) is identical to the `bos_token_id` (50256), `eos_token_id` (50256), or the `sep_token_id` (None), and your input is not padded.
Evaluating: 26it [00:00, 75.66it/s]

Evaluation Results:
  - Average NLL         : 4.2212
  - Bits Per Token (BPT): 6.0899
  - Perplexity (PPL)    : 68.1135
  - Total Tokens        : 9782
  - Model Time (s)      : 0.19
  - Speed (batches/sec) : 135.85
\end{verbatim}

\begin{verbatim}
Sample generations from model (unprompted):

Generated Paragraph 1:
 = formalist , liberal , moderate , conservative , and democratic . He was an active member of the Social Democratic Party ( SN ) for a total term of five years from 1949 to 1951 ; he served on its executive committee from 1956 until his death in 1968 at age 78 . In 1962 – 64 , he led the party's campaign against racial segregation as part one of their " national program " . Although he left the party after his election , he continued running for re - election in 1966 when it won two seats in the New Hampshire House : the seat that had been vacated by Thomas W. Hartigan in 1963 . After retiring as Governor General of Massachusetts in 1965 , Stasi worked as a political consultant and fundraiser . He died on April 17 , 2008 at the age 81 . 
 Featured articles : The Harvard Crimson review 
Online archive of all his papers 

Neue Suddeutsche Zeitung website  www.neuwjhts
--------------------------------------------------------------------------------

Generated Paragraph 2:
 = Subsequent releases and remixes , including : Drowned in Sound ( 2008 ) ; The One I Love , featuring Jay Z , with Jay - Z and Kanye West . It 's All Too Much , Part II ( 2010 – 2011 ) . 
BeATBEAM was released as a double A & R single in Europe on July 26 , 2011 , in the United States , on August 24 , 2011 through iTunes Store . Bearer also made his second appearance on September 16 at the 2015 Grammy Awards and was nominated for Best Rap Performance by an African American Artist for " Born This Way " . 
 
 = Later appearances 
formally included as one of four singles from the album , along with " Say You 'll Still Love Me " , which was released to promote it .  Also performed were two songs from the extended play , titled " Not Myself Tonight " , and three additional tracks , entitled " What 's Going On Now ?
--------------------------------------------------------------------------------

Generated Paragraph 3:
 = 
 
 = The two @-@ part episode " A New Day in the Life of John Lennon and McCartney " was broadcast on BBC One , BBC Two HD ( with a repeat run of 12 minutes ) . 
formats : widescreen , Dolby Digital 5 @.@ 1 Surround soundtrack 
 ' s audio track is available in MP3 / FLAC format 
is available in stereo or mono 
and can also be played at any time . Note that this music video was not edited to make it appear as if the entire show were filmed in London .  There are no recordings from the second series shown . In addition there appears an uncredited cameo appearance by actor Hugh Laurie in his role as Professor John Watson , who is seen dancing with Peter after the last meeting .  It has been suggested that the scene where Pete meets John when he talks about his work as a journalist had originally appeared in episodes such as " A Town Called Mercy "
--------------------------------------------------------------------------------
(/local/data1/thawe276/ENVautoregressive) lnx00293 [/local/data2/home/thawe276/Documents]> 
\end{verbatim}

\section{D3PM Model Outputs}
\label{d3pm:d3pm-images}

\begin{verbatim}
--------------------------------------------------------------------------------------------------------------------
<|endoftext|> gold by the Australian Recording Industry Association" leading to re-entering the second international single, Blood on the Ice, which stayed inside the top for ten weeks. The single peaked at number two and at five while reaching a peak of just under two weeks. The song was later given a gold certification by the Australian Recording Industry Association (ARIA), denoting shipments of 500,000 physical units.
<|endoftext|><|endoftext|> === Singles ===
<|endoftext|><|endoftext|> The lead single from The Dual was "My Love Remains", which was produced and officially released on the Billboard Singles Chart at number 38, two weeks later. It was his fastest debut in the United States, reaching number 10 and spending 10 weeks on the chart, peaking off "4 Minutes", at 68 on the artist's chart and the song's peak at 11, it peaked at number eight of the Billboard Hot 100. The single received positive feedback from music critics, who praised the song's twist and sound, and his revealing voice as loud and smooth. "I Want" earned Nelly a Grammy Award nomination for the Best Vocal Performance in 1988 while Nelly performed on two other dates. The live performance was announced as a small party with rapper Fewly doing choreography to front of the audience. This live performance was noted for Furt playing him's most obscure song on the album, "All I Ever."
<|endoftext|> In February 1989, it reached the number nine on the Billboard Hot 100 for 20 weeks. In May 1989, the song peaked at number two on the US Billboard Hot 100 — becoming his first entry on the chart — behind Johnny Cash's "Not Adjustment Season". It eventually peaked at number 24 on the Billboard Top Digital Songs chart, and became Rocky's first single to top the Hot 100. The song alone peaked at number two each during four other regional hit in the charts. By year 1989, "She" reached the top of the US Billboard Hot 100. By 1990, it had shipped 2.6 million copies in the United States. After six weeks, the single had been certified double platinum by the now-RIAA.
<|endoftext|> "I Like a Woman" had also a chart success and entered the Australian charts, reaching number twenty-one. The song peaked at number one on the chart for one week, his most successful single of 1989, denoting a top 40 of 12 weeks from the album. Like only Live in a European continent, it reached a peak at number one, where "Without You", sold by the Australian Recording<|endoftext|>


\end{verbatim}

\begin{verbatim}
<|endoftext|> short film, which featured drums to Al Savage, compared it to Ray Charles'" heavy-fast and "Bad Boy", which both reached the Hot 100 singles chart. At that time, bassist Bob Frearsud played backing, and pianist Harrison gave in his vocals.
<|endoftext|><|endoftext|> While working for the initial stages of his recording career, Yankovic began to shy away from his musical direction for himself. In an interview with Rolling Stone in 1992, Yankovic mentioned that the sessions as "My first you say,' I don 't think about the possibility of the change in my life [family]..'" As with the progress stages of making a debut album, Yankovic considered ", 'Well my first guy did come ready for music.' I am not second and then make it feel the whole?' I did not know why I was doing when there was ever something that was trying to be right on my own. That was a personal thing telling me to think I did things in a lot of that work [...] So I just like myself different works. '
<|endoftext|> In 2002, the compilation album included four new songs. Yankovic had asked to have songs that Yankovic wrote for the compilation album, "This Love", "funny", "Off the Teal Vah Neck" and "Weird Al Part II were finished by October 2002.
<|endoftext|> In 2001, Yankovic released "This Art Don 't" (Remixes) together with Yankovic recording ten new songs. It marked Yankovic's 30th studio album, entitled Permanent Record 11: 9 (2003). A CD reissue entitled It's A Liar It featuring Jay Lucas was released between August 1999 and February 2003; and "NizYoterPresking" was released on November 27, 2000. The CD's release in the United Kingdom on April 10, 2002 included the single "(Last Gought I'm Time)" and "My Heart". 1993's The Girl Hotel Italy reached number 20 in the UK Singles Chart. "Kiss Me Out" reached its peak in the Top 10 in 1990. The album became his second Top 50 hits in 1990 and the United Kingdom. The same year, he performed a television medley of My Sun, and the 2007a Ray Charles tribute show Before Getting Out of the album. The series was nominated for both Best Overall Album, and Pacific ASCAP: American Albumists of the Year.
<|endoftext|> Yankovic had not been with the studio for 10 years since<|endoftext|>

\end{verbatim}

\begin{verbatim}
--------------------------------------------------------------------------------------------------------------------
<|endoftext|> their own state when Ohio got elected. Although McKinley was doubtful of strong views to run his political career, and McKinley received that they should be the right to become governor on a titular basis, the governor urged them to fight what they see in Ohio to name Bryan, and McKinley agreed to campaign for the presidency.
<|endoftext|> Newspapers in New Jersey called the McKinley election even the best, the state newspapers deemed him scandalous and political bosses went back to California to cover him. A close friend, after a visit of Long Bryan attacked McKinley, stating that McKinley, having run, did not seek the governor's leadership it would be his business to stay away, and was nominated by the state's alumnus. McKinley wrote to Governor General MacArthur, "And that come is to ask the press he 'd done so much else, and, if he had borrowed any money, and say why, take the general, that man did not insult me, let indeed let me go.", after having seen McKinley at the hotel after holding the governor away with his apology, McKinley turned Bradley down the nomination, and wrote to Reed to appoint him their lieutenant governor. New York governor Bradley Jr refused the appointment — "while he preferred McKinley to have employed them ... which lasted six months, but now President Harrison would have. Or this would have favored them until the plaud was put in place ... I thought,' do I not, did you have the discussion? '" The former Ohio governor thought McKinley had received the entreaties and taken over the Senate seat, showing McKinley as "the advancement and prosperity of the United States, the leader of a nation without history, and in a time when American democracy and free will were wiped out in its years of existence. The highest triumph was to government by McKinley." The influence of industrialism made the governor's name apparent in McKinley and his later life.
<|endoftext|><|endoftext|> == Issue ==
<|endoftext|><|endoftext|><|endoftext|><|endoftext|> = Thated dollar (18cent) =
<|endoftext|><|endoftext|> The second coin is the type of three-cent design; it is a combination of silver, half dollar, and eagle designs designed by Chief Engraver James Longacre, and bears his original design. Longacre's first coin design for the first year (1804) was the theacre dollar, and at the request of the Mint on both sides the half eagles were designed under the rule of John C. Patterson.
<|endoftext|> Longacre in 1804<|endoftext|>


\end{verbatim}

\begin{verbatim}
--------------------------------------------------------------------------------------------------------------------
python ./mdlm/main.py
    model=small\ 
    data=wikitext103\ 
    parameterization=d3pm\
    model.length=512\
    eval.compute_generative_perplexity=True \
    sampling.steps=1000 \
    sampling.predictor=ddpm \
    loader.batch_size=8 \
    loader.eval_batch_size=4 \
    time_conditioning=True data.cache_dir=/local/data2/home/thawe276/.cache/textdiffusion \
    seed=1000 \
    T=1000


\end{verbatim}
	
	
	

\end{document}